\documentclass[lettersize,journal]{IEEEtran}
\usepackage{amsmath,amsfonts}
\usepackage{algorithmic}
\usepackage{algorithm}
\usepackage{array}
\usepackage[caption=false,font=normalsize,labelfont=sf,textfont=sf]{subfig}
\usepackage{textcomp}
\usepackage{stfloats}
\usepackage{url}
\usepackage{verbatim}
\usepackage{graphicx}
\usepackage{cite}
\usepackage{xcolor}
\usepackage{booktabs}  
\usepackage{multirow}  
\usepackage{lipsum}
\usepackage{bbding}
\usepackage{bm} 
\usepackage{mathrsfs} 
\usepackage{graphicx}
\usepackage{xcolor}
\usepackage{booktabs}
\usepackage{rotating}
\usepackage{amssymb}

\usepackage[colorlinks,
            linkcolor=red,
            anchorcolor=blue,
            citecolor=green
            ]{hyperref}

\begin{document}
\title{GMG: A Video Prediction Method Based on Global Focus and Motion Guided}

\author{Yuhao Du, Hui Liu, Haoxiang Peng, Xinyuan Cheng, Chengrong Wu, Jiankai Zhang$^{*}$ \thanks{* Corresponding author}
\thanks{This research was jointly supported by the National Natural Science Foundation of China (U2442211), Scientific Research Fund of Hunan Provincial Education Department (24C0128). The computations in this research was supported by Supercomputing Center of Lanzhou University. The source code will be released at \href{https://github.com/duyhlzu/GMG}{https://github.com/duyhlzu/GMG}.}
\thanks{Yuhao Du and Jiankai Zhang are with the College of Atmospheric Sciences, Lanzhou University, Lanzhou 730000, China (e-mail: duyh2024@lzu.edu.cn; jkzhang@lzu.edu.cn).}% <-this % stops a space
\thanks{Hui Liu and Haoxiang Peng are with the College of Computer and Mathematics, Central South University of Forestry and Technology, Changsha 410004, China (e-mail: T20030970@csuft.edu.cn; 20235223@csuft.edu.cn).}
\thanks{Xinyuan Cheng is with the Center for Language and Information Processing, University of Munich (LMU), Munich 80539, Germany (e-mail: Xinyuan.Cheng@campus.lmu.de).}
\thanks{Chengrong Wu is with the Department of Computer Science, University of Manchester, Manchester M139PL, UK (e-mail: chengrong.wu@student.Manchester.ac.uk).}
}

% The paper headers
\markboth{Journal of \LaTeX\ Class Files,~Vol.~14, No.~8, August~2021}%
{Shell \MakeLowercase{\textit{et al.}}: A Sample Article Using IEEEtran.cls for IEEE Journals}

%\IEEEpubid{\begin{minipage}{\textwidth}\ \centering
%		Copyright © 2026 IEEE. Personal use of this material is permitted. \\
%		However, permission to use this material for any other purposes must be obtained from the IEEE by sending an email to pubs-permissions@ieee.org.
%\end{minipage}}

\maketitle

\begin{abstract}
Recent years, video prediction has gained significant attention particularly in weather forecasting. However, accurately predicting weather remains a challenge due to the rapid variability of meteorological data and potential teleconnections. Current spatiotemporal forecasting models primarily rely on convolution operations or sliding windows for feature extraction. These methods are limited by the size of the convolutional kernel or sliding window, making it difficult to capture and identify potential teleconnection features in meteorological data. Additionally, weather data often involve non-rigid bodies, whose motion processes are accompanied by unpredictable deformations, further complicating the forecasting task. In this paper, we propose the GMG model to address these two core challenges. The Global Focus Module, a key component of our model, enhances the global receptive field, while the Motion Guided Module adapts to the growth or dissipation processes of non-rigid bodies. Through extensive evaluations, our method demonstrates competitive performance across various complex tasks, providing a novel approach to improving the predictive accuracy of complex spatiotemporal data.
\end{abstract}


\section{Introduction}

\IEEEPARstart{I}{n} recent years, spatiotemporal prediction has garnered unprecedented attention due to the growing demand for prediction tasks across various domains, particularly in weather forecasting \cite{b1,b2,b37} and traffic prediction \cite{b3,b4}. Accurate predictions can bring significant social and economic value. Currently, spatiotemporal prediction models based on deep learning method can be categorized into four main types: RNN, CNN, CNN-RNN, and ViT.

RNN models primarily rely on stacking basic predictive units to transmit information in both depth and breadth. The advantage of RNN lies in its ability to capture long-term temporal features, though these models generally involve a large number of parameters. Classic RNN models include PredRNN \cite{b4}, PredRNN++ \cite{b5}, MIM \cite{b6}, MotionRNN \cite{b7}, PredRNN-V2 \cite{b8}, and VMRNN \cite{b9}. These models improve prediction performance by optimizing the organization and transmission of information flows or parameter handling.

CNN models focus on capturing multiscale features through techniques such as multiscale convolution stacking, upsampling, and skip connections. The strength of CNN models lies in their simplicity, allowing them to capture multiscale features efficiently, although they often struggle to capture long-term temporal features. Notable CNN models include PredCNN \cite{b10}, SimVP \cite{b11}, TAU \cite{b12} and  TrajectoryCNN \cite{b34}, which have demonstrated excellent performance and interpretability using streamlined predictive frameworks, offering diverse solutions for spatiotemporal sequence prediction.

Hybrid CNN-RNN models enhance the predictive capability of RNNs by preprocessing input data and extracting multiscale features using CNNs. Models like E3D-LSTM \cite{b13}, CrevNet \cite{b14}, and PhyDNet \cite{b15} achieve accurate and efficient predictions through different CNN encoders for effective feature extraction.

ViT primarily utilizes the Transformer architecture, which has gained popularity in recent years, leveraging the attention mechanism to capture image features. SwinLSTM \cite{b16}, SimVP-ViT \cite{b17} and PredFormer \cite{b18} are built on Vision Transformer \cite{b19} or Swin Transformer \cite{b20} backbones. As an emerging approach, ViT models have been proven to possess strong predictive capabilities. However, due to the high computational complexity of attention mechanisms, they typically require more training resources.

\begin{figure}[htbp]
\centerline{\includegraphics[width=8cm]{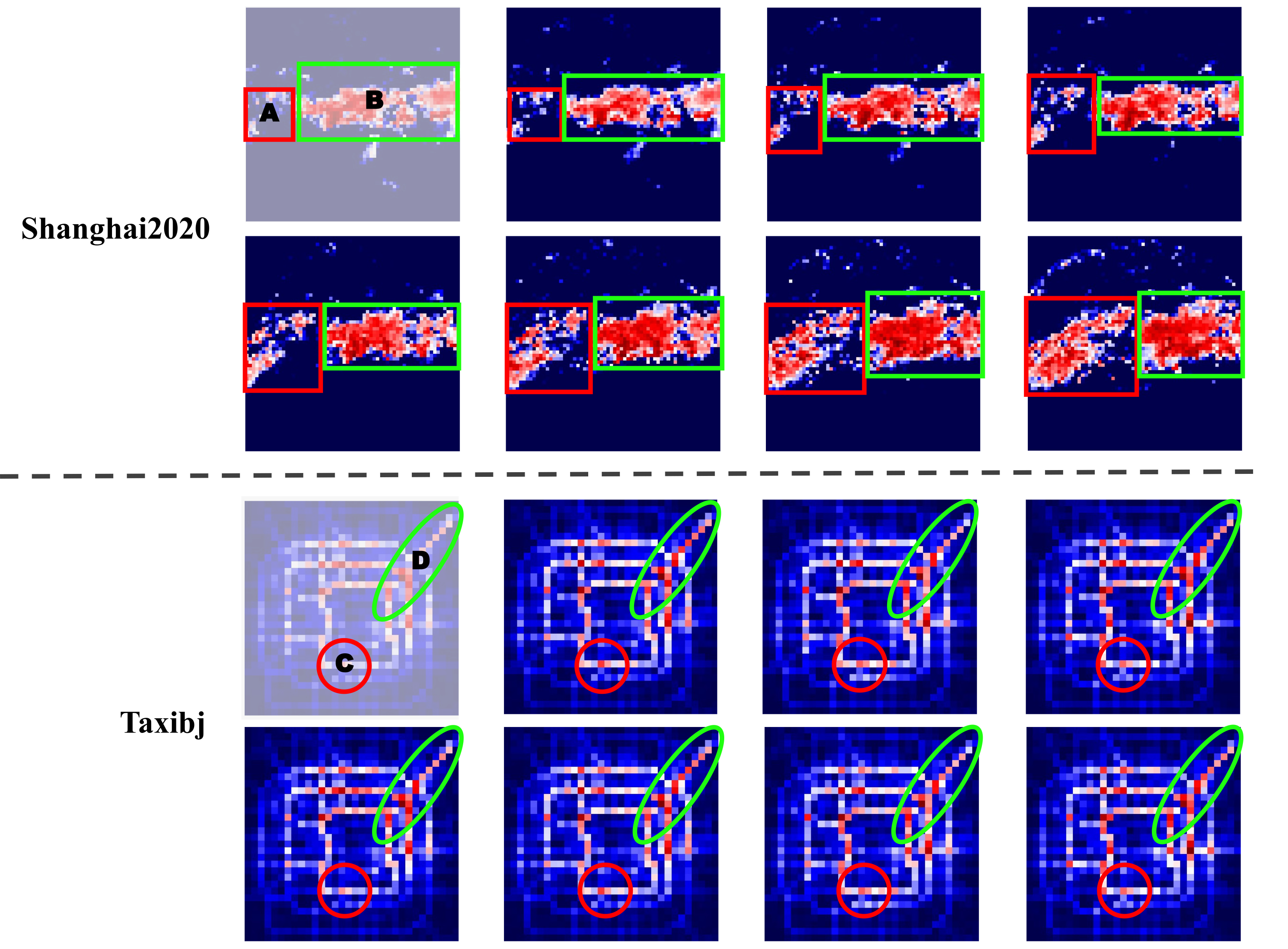}}
\caption{The two main real-world challenges addressed in this paper. Position and shape changes of rainfall regions over time: For example, in region B, while moving eastward, the north-south extent of the rainfall gradually expands, adding complexity to prediction tasks. Long-range correlations: In real-life scenarios, a traffic jam in region C may be related to traffic in another part of the city, such as region D. These long-range correlations are often difficult to capture, limiting the accuracy of predictions for such data.}
\label{fig:one}
\end{figure}

Given its timeliness and societal relevance, meteorological data prediction is of great research interest. Yet, spatiotemporal models, particularly convolution-based ones for feature learning, encounter two challenges (Fig. \ref{fig:one}): (1) Localized feature extraction limits the ability to capture long-range dependencies, crucial for modeling large-scale meteorological patterns and non-local effects such as teleconnections. (2) The non-rigid nature of atmospheric elements (e.g., cloud formations and storm systems) makes it difficult to accurately model their evolution over time. 

Taking precipitation data as an example, rain clouds not only shift in a specific direction but also undergo overall deformation. The dissipation and growth of clouds over time are particularly difficult to predict accurately. In addition, in real life, a traffic jam on one street may impact another distant street. In meteorological data, analogous phenomena exist where climate anomalies exhibit correlations over long \IEEEpubidadjcol distances. 
However, the complexity and unpredictability of such long-range correlations significantly limit prediction accuracy.

Previous studies have developed models that decouple motion patterns to better predict object motion. MotionRNN, for instance, decomposes motion into transient variation and trending momentum to model object motion dynamics. MAU \cite{b21} separately trains augmented motion information (AMI) and current spatial state to capture reliable inter-frame motion information. MMVP \cite{b22} constructs an appearance-independent motion matrix to decouple motion and appearance information, inferring future object motion from image frames while maintaining appearance consistency across frames. MAFE \cite{b32} proposes a Motion Perceptual Loss and extracts object-perceptive features through an attention module. HO-LMC \cite{b33} proposes a high-order motion encoder to capture high-order motion dynamics in image sequences. These studies provide valuable theoretical foundations. However, there is still a lack of effective feature extraction methods for non-rigid object motion.

Inspired by non-rigid motion modeling, this paper introduces a Motion Guided Module (MGM). Building upon conventional motion-state decomposition, MGM incorporates two deformation factors—balance factor $\alpha$ and decay factor $\beta$—to characterize local growth and dissipation, global deformation, and inelastic morphological changes of a target during motion. The balance factor regulates the contribution of local motion to the overall shape, ensuring a spatially coherent distribution of motion information. The attenuation factor models the temporal decay or dissipation of morphology, thereby enabling more precise prediction of dynamic changes in non-rigid targets. Through this design, MGM captures fine-grained features of complex target motions and improves the prediction accuracy of non-rigid motion sequences.

Meanwhile, existing approaches extend a model's global receptive field through enhanced information-flow mechanisms, multi-scale convolutional stacks, or sliding-window strategies. However, a key challenge remains: whether based on convolution or sliding windows, the receptive field is inherently limited, making it difficult for models to fully capture and generalize global features when dealing with spatiotemporal data with potential long-range dependencies. To address this issue, this paper proposes a Global Focus Module (GFM), a non-attention, low-complexity global feature aggregation method. By extracting global features from the input image \(X\) and integrating them with the hidden state \(H\), GFM substantially enhances the model's global perspective. Compared with traditional attention-based global modeling methods, GFM can capture latent long-range dependencies without significantly increasing computational complexity, thereby improving the model's predictive capability on large-scale spatiotemporal sequences.

The proposed GMG (\textbf{G}lobal focus and \textbf{M}otion \textbf{G}uided) predictive model consists of four main components: SpatioTemporal ConvLSTM (ST-ConvLSTM), Motion Guided Module (MGM), Self-Attention Memory, and Global Focus Module (GFM). We aim to offer a general solution for video prediction, particularly for meteorological data, and provide a novel approach to achieving accurate meteorological forecasting. The main contributions of this paper are summarized as follows:

\begin{itemize}
    \item We propose a more comprehensive object motion decoupling method, the \textbf{Motion Guided Module (MGM)}, which captures non-rigid deformation characteristics by introducing a \textbf{balance factor} and a \textbf{decay factor}.
    \item We develop a \textbf{Global Focus Module (GFM)} to address the limitations of convolutional receptive field size, enabling the model to capture potential long-range feature correlations.
    \item By integrating MGM and GFM, we propose the \textbf{GMG video prediction framework}. Extensive experiments on six different types of datasets demonstrate that GMG achieves state-of-the-art performance, providing an innovative solution for video prediction.
\end{itemize}

\section{Related Work}
Over the past decade, spatiotemporal prediction using recurrent structures has originated from ConvLSTM, a simple yet effective framework that remains widely used today. Following this, numerous RNN-based models have been proposed. PredRNN \cite{b4} introduced ST-ConvLSTM, which added a Spatiotemporal Memory module parallel to the cell state in ConvLSTM and organized a cross-layer information flow. This design allowed the model's final output to comprehensively capture hierarchical spatiotemporal features. Building upon PredRNN, PredRNN++ \cite{b5} regulated the sequence of spatiotemporal memory and cell states, leading to the formation of Causal LSTM. Additionally, PredRNN++ incorporates the Gradient Highway Unit (GHU), which facilitates fast information flow between adjacent recurrent units and effectively mitigates the vanishing gradient problem along the temporal dimension.

To further enhance the prediction capability for non-stationary data, MIM introduced a differential computation between the hidden states of adjacent recurrent nodes, effectively reducing the higher-order non-stationarity in spatiotemporal data. MIM also reorganized the hidden state transmission mechanism, enabling diagonal transmission of information. This allowed basic units to fully utilize the differential pattern. MotionRNN \cite{b7} extended PredRNN by decomposing hidden states into two motion states: Transient Variation ($F$) and Trending Momentum ($D$), trained using GRU. This approach achieved accurate predictions for moving objects and demonstrated strong performance on radar echo data. SwinLSTM \cite{b16}, inspired by Swin Transformer, segmented images into patches as tokens for input while retaining the fundamental RNN framework, introducing a new perspective for RNN-based improvements. Similarly, VMRNN \cite{b9} combined Vision Mamba with RNN. Like SwinLSTM, it processed image tokens but employed a selective state-space model for input handling.

Notably, the GMG proposed in this paper is also implemented based on the fundamental RNN architecture as a basic predictive unit. Furthermore, the core module of GMG, the Global Focus Module, is a flexible and independent component. This design allows it to integrate seamlessly with the aforementioned RNN-based predictive models without structural modifications. Unlike previous prediction methods, we emphasize expanding the model's global perspective and designing a learning module specifically for handling shape variations in moving objects. Our approach provides a novel solution for video prediction, particularly for meteorological data with complex dynamic processes.

\begin{figure*}[htbp]
\centerline{\includegraphics[width=20cm]{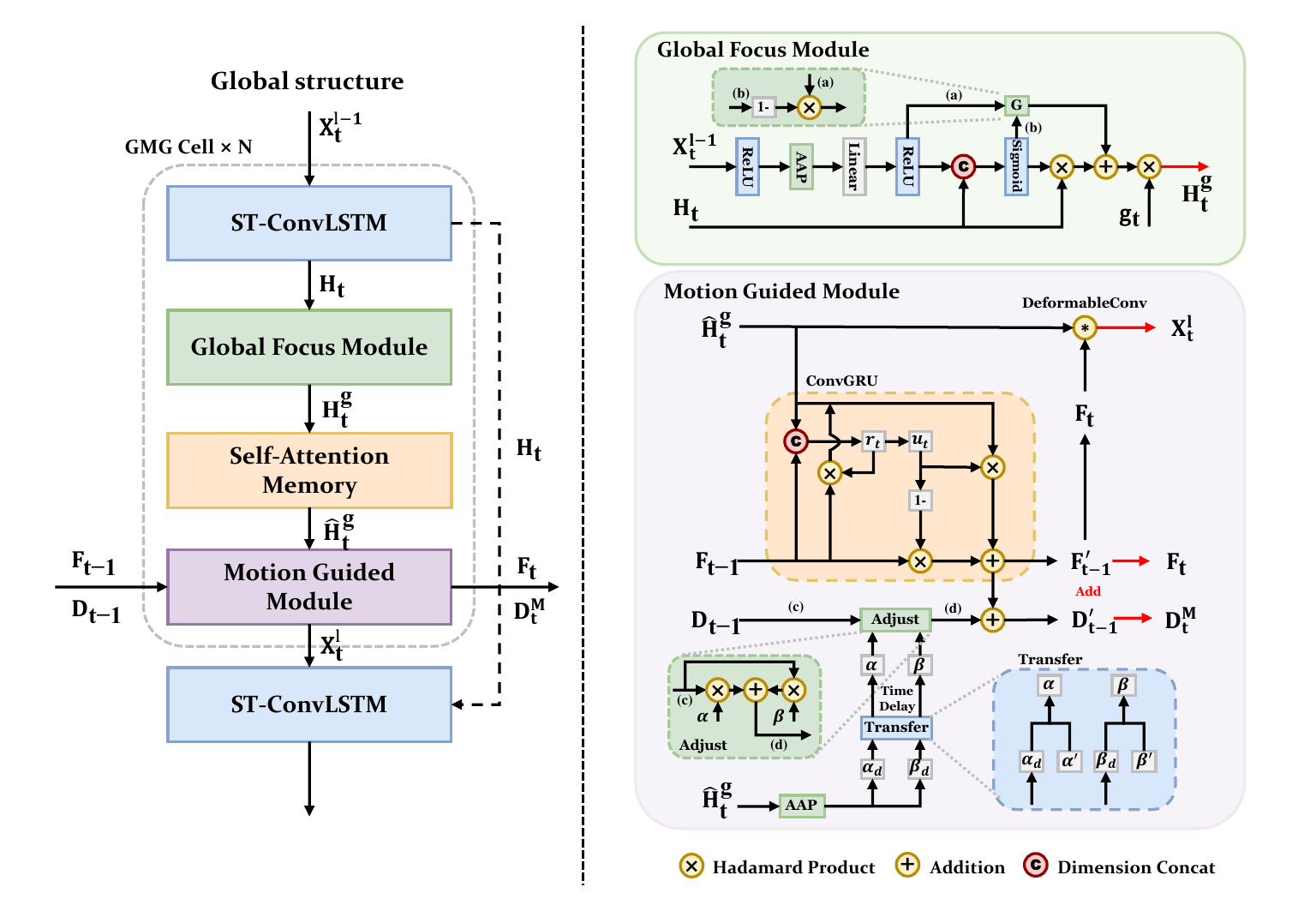}}
\caption{The overall structure of GMG (left) and the two main modules proposed in this paper (right) are illustrated. A standard GMG unit consists of four modules: ST-ConvLSTM, Global Focus Module, Self-Attention Memory, and Motion Guided Module. In this study, each time step is composed of four stacked GMG units. The temporal memory \( M \) from the fourth layer at time \( t \) is transferred to the first layer at time \( t+1 \), ensuring that the model captures long-term temporal dynamics effectively. The "Time Delay" in the figure refers to the operation corresponding to Eq.(\ref{eq:fifteen}) in the text.
}
\label{fig:two}
\end{figure*}

\section{Methods}

In this section, we introduce the proposed GMG prediction model. As mentioned above, our model is designed to address two key challenges in spatiotemporal forecasting:

(1) How to effectively capture non-rigid motion deformations to improve the accuracy of object movement prediction?

(2) How to enhance the model's ability to extract global dependencies in order to better capture long-range correlations (eg., teleconnections) in spatiotemporal data?

In the following subsections, we first define the problem and then introduce in detail how our GMG model addresses the two aforementioned challenges.

\subsection{Problem Definition}\label{A}
The video prediction problem can be formulated as follows: Given a sequence of consecutive images \( X \) over a certain period, the goal is to find a set of parameters \( \theta^* \) that maximize the probability of the predicted image sequence \( Y \):

\begin{equation}
\theta^* = \arg\max_{\theta} P(Y | X; \theta).
\end{equation}

where \( \theta \) represents the model parameters to be trained.

In this work, we adopt the \( L_2 \) loss for optimization, which assumes the prediction error follows a Gaussian distribution. Under this assumption, maximizing the likelihood is equivalent to minimizing the following loss:

\begin{equation}
\mathcal{L}_{L_2} = \sum_{i} (Y_i - \hat{Y}_i)^2,
\end{equation}

where \( \hat{Y} = F_\theta(X) \) represents the predicted future frames generated by the model \( F_\theta \) given the input sequence \( X \). The \( L_2 \) loss penalizes larger deviations more strongly, encouraging smooth and consistent predictions.

\subsection{Overall Architecture}
The overall architecture of GMG is presented on the left side of Fig. \ref{fig:two}. The input image \( X \) is processed through ST-ConvLSTM to update the hidden state \( H_t \). Subsequently, \( H_t \) is fed into the Global Focus Module (GFM), where it is fused with global features to produce \( H_t^g \), enabling the model to perceive global characteristics of the input data. The primary role of Self Attention Memory (SAM) from SA-ConvLSTM \cite{b24} is to integrate \( H_t^g \) with the spatiotemporal memory \( M_t \) to capture spatial contextual information, thereby enhancing the model's ability to maintain long-term feature dependencies. Finally, \( \hat{H}_t^g \) is input into the Motion Guided Module (MGM), which regulates the model's motion state learning ability through two state factors, \( \alpha \) and \( \beta \). GMG retains the cross-layer transmission of spatiotemporal memory and the Gradient Highway Unit \cite{b5}, ensuring effective utilization of various types of valuable information.

\subsection{Global Focus Module}

In Section 1, we discuss the challenges that existing methods face in capturing long-range correlations. In this section, we introduce a global correction approach: the Global Focus Module  (Fig. \ref{fig:two}, right). In the actual training process, features are typically not extracted from the entire image but are processed by dividing the image into patches or using a sliding window approach. Additionally, due to the limitation of the kernel size, the receptive field of the model is restricted, which leads to the model's inability to capture potential correlations between distant regions in the image. Therefore, we propose a visual focusing approach. After processing the local features, we revisit the entire image's features and inform the model of the important global region correlations to capture the potential long-range dependencies.

First, we apply Adaptive Average Pooling (AAP) to extract the \textbf{Global Features} (\( \bm{G_{F}} \)) from the input image \( \bm{X_{t}} \in \mathbb{R}^{T \times C \times H \times W} \):
\begin{equation}
\bm{G_{F}} = \mathcal{AAP}(\texttt{ReLU}(\texttt{Conv2d}(\bm{X_t}))).
\end{equation}

Adaptive Average Pooling (AAP) divides the input feature map \( X_t \) into rectangular regions and outputs the average value of all elements within each region. Formally, for the \( i \)-th feature map and region \( R_j \):

\begin{equation}
\mathcal{AAP}(\bm{X_t})_{i, j} = \frac{1}{|R_j|} \sum_{k \in R_j} \bm{X_t}^{(i)}(k)
\end{equation}

where \( \bm{X_t}^{(i)}(k) \) is the element at position \( k \) in region \( R_j \) of the \( i \)-th feature map, \( AAP(X_t)_{i,j} \) denotes the result of average pooling over the \( j \)-th region of the \( i \)-th channel feature map at time \( t \). \( |R_j| \) is the number of elements in \( R_j \). In the actual implementation, the output size is set to \( 1 \times 1\), which \( \bm{G_{F}} \in \mathbb{R}^{T \times C \times 1 \times 1} \).

Then, a linear fully connected layer is used to map the \( \bm{G_{F}} \) to a specified dimensional space:

\begin{equation}
\bm{X_t^g} = \texttt{ReLU}(\texttt{Linear}(Enc(\bm{G_{F}}))).
\end{equation}
where, $Enc(\cdot)$ maps $\bm{G_{F}}$ from the shape $\mathbb{R}^{T \times C \times 1 \times 1}$ to $\mathbb{R}^{T \times C \times H \times W}$.

We also design a gate mechanism to control the degree of fusion between the global features and the hidden state:

\begin{equation}
\mathcal{\bm{G}} = \texttt{Sigmoid}(\texttt{Conv2d}_{W_{\mathcal{\bm{G}}}}(\texttt{Concat}[\bm{H_t},\bm{ X_t^g}])),
\end{equation}
\begin{equation}
\bm{H_t^g} = \bm{H_t} \circ \mathcal{\bm{G}} + \bm{X_t^g} \circ (\bm{1} - \mathcal{\bm{G}}).
\end{equation}

Where, \(\bm{X_t^g},\bm{H_t} \in \mathbb{R}^{T \times C \times H \times W} \), \( W_{\mathcal{\bm{G}}}\) represents convolution kernel, \( \circ \) denote the Hadamard product. To ensure that the updated hidden state includes more global information across different scales, we apply a multi-scale focus method:

\begin{equation}
\bm{g_t} = \texttt{MultiScaleConv}_{(1,3,5)}(\bm{X_t}),
\end{equation}
\begin{equation}
\bm{H_t^g} = \bm{H_t^g} \circ \bm{g_t} \label{eq:six}.
\end{equation}

Where \(\texttt{MultiScaleConv}_{(\ast)}\) represents feature extraction using different-sized convolution kernels, and the final output is the sum of all feature maps, \( \bm{g_t} \in \mathbb{R}^{T \times C}  \). Then, \( \bm{g_t} \) will be broadcasted to the same shape as \( \bm{H_t^g} \in \mathbb{R}^{T \times C \times H \times W} \). We refer to the operation in Eq.(\ref{eq:six}) as "Feature Focus".

\subsection{Self-Attention Memory}

To further enhance the hidden state $\bm{H_t^g}$ after global feature correction, we introduce the \textbf{Self-Attention Memory (SAM)} module, which captures the relationship between the current hidden state $\bm{H_t^g}$ and the long-term memory $\bm{M_{t-1}}$ through a self-attention mechanism.

The SAM takes two inputs: the globally enhanced hidden state $\bm{H_t^g} \in \mathbb{R}^{T \times C \times H \times W}$ and the memory from the previous timestep $\bm{M_{t-1}} \in \mathbb{R}^{T \times C \times H \times W}$. To facilitate attention computation, we first flatten the spatial dimensions into a sequence. Let $N = H \times W$, and reshape the inputs as $\bm{H_t^g}, \bm{M_{t-1}} \in \mathbb{R}^{T \times C \times N}$.

Next, the hidden state $\bm{H_t^g}$ is linearly projected to obtain the query $\bm{Q_h} \in \mathbb{R}^{T \times d \times N}$, and the memory $\bm{M_{t-1}}$ is projected to generate the key and value, $\bm{K_m}, \bm{V_m} \in \mathbb{R}^{T \times d \times N}$, where $d$ is the attention embedding dimension. The similarity score between query and key is computed as:

\begin{equation}
\bm{e_m} = \bm{Q_h}^\top \bm{K_m} \in \mathbb{R}^{T \times N \times N},
\end{equation}

This is followed by a softmax operation to obtain normalized attention weights:

\begin{equation}
s_{m;i,j} = \frac{\exp(e_{m;i,j})}{\sum_{k=1}^{N} \exp(e_{m;i,k})},
\end{equation}

The memory features are then aggregated using these weights:

\begin{equation}
\bm{Z_{m;i}} = \sum_{j=1}^{N} s_{m;i,j} \cdot \bm{V_{m;j}} = \sum_{j=1}^{N} s_{m;i,j} \cdot \bm{W_{mv}} \bm{M_{t-1;j}},
\end{equation}

where $\bm{W_{mv}}$ is a learnable projection matrix, and $\bm{M_{t-1;j}}$ denotes the $j$-th spatial location of the memory.

Finally, the memory-based output $\bm{Z_m}$ is concatenated with a self-attention output $\bm{Z_h}$ computed from $\bm{H_t^g}$, and the combined result is fused via a linear projection:

\begin{equation}
\bm{Z} = \bm{W_z}[\bm{Z_h}; \bm{Z_m}],
\end{equation}

where $\bm{W_z}$ is a learnable fusion matrix. This fused representation $\bm{Z}$ is then passed to a gated memory update mechanism described in the following section.

To maintain long-term consistency, the memory is updated through a gated mechanism inspired by LSTM, but optimized for efficiency. Instead of full convolution, we use \textbf{depthwise separable convolutions}~\cite{b36} to reduce parameters:
\begin{align}
\bm{i_t^{a}} &= \sigma({W_{zi}} * \bm{Z} + {W_{hi}} * \bm{H_t^g} + \bm{b_i}^{a}), \\
\bm{C_t^{a}} &= \tanh({W_{zc}} * \bm{Z} + {W_{hc}} * \bm{H_t^g} + \bm{b_c}^{a}), \\
\bm{M_t} &= (1 - \bm{i_t^{a}}) \circ \bm{M_{t-1}} + \bm{i_t^{a}} \circ \bm{C_t^{a}}.
\end{align}

The final output hidden state $\hat{\bm{H_t}}$ is modulated by an output gate:
\begin{align}
\bm{o_t^{a}} &= \sigma({W_{zo}} * \bm{Z} + {W_{ho}} * \bm{H_t^g} + \bm{b_o^{a}}), \\
\bm{\hat{{H_t^{g}}}} &= \bm{o_t^{a}} \circ \bm{M_t}.
\end{align}

Overall, the SAM module enables efficient retrieval and integration of historical contextual information through attention-weighted memory and lightweight gating mechanisms. Its use of depthwise separable convolutions makes it suitable for deployment in high-resolution scenarios.

It is worth noting that SAM and GFM are complementary. Unlike conventional global attention, GFM does not compute pairwise attention weights across regions; instead, it first extracts global statistical features via adaptive pooling and then applies gated multi-scale convolutions to globally calibrate the hidden state, thereby explicitly introducing long-range dependencies and cross-region correlations while avoiding high computational cost and memory overhead. Building on this, SAM retrieves and integrates long-term memory on demand along the temporal dimension, performing its self-attention operation only on compressed memory representations and employing depthwise separable convolutions for efficient gated updates. Consequently, the combination of GFM and SAM rapidly focuses on global spatial context while efficiently leveraging historical information over time, enabling richer and more stable spatiotemporal representations at lower complexity than global attention mechanisms—particularly well suited for video prediction tasks involving complex dynamics and long time horizons.

\subsection{Motion Guided Module}
Inspired by the motion decomposition in MotionGRU \cite{b7}, we propose a decomposition model based on motion evolution. This model divides the spatial movement of objects into transient variations \( \mathcal{F} \) and trending momentum \( \mathcal{D} \). \( \mathcal{F}, \mathcal{D} \in \mathbb{R}^{T \times 2k^2 \times \frac{H}{2} \times \frac{W}{2}} \), \(k\) is filter size. Additionally, MGM introduces for the first time in video prediction two physically interpretable deformation factors—the balance factor and the decay factor (Fig. \ref{fig:two}). The balance factor models the “growth trend” of motion, reflecting a target’s tendency to preserve momentum and internal deformation during movement, thereby allowing the model to account for both instantaneous changes and historical motion states during prediction. The decay factor simulates the temporal dissipation of motion energy, applying an exponential decay to motion trends to suppress the influence of outdated dynamics, thus enhancing stability when handling long video sequences.

First, the MGM processes the transient variations \( F \) and trending momentum \( D \) using GRU:
\begin{equation}
\bm{u_t} = \sigma(\texttt{Conv2d}_{W_{\mathcal{\bm{U}}}}(\texttt{Concat}([\bm{\hat{{H_t^{g}}}}, \mathcal{F}_{t-1}]))),
\end{equation}
\begin{equation}
\bm{r_t} = \sigma(\texttt{Conv2d}_{W_{\mathcal{\bm{R}}}}(\texttt{Concat}([\bm{\hat{{H_t^{g}}}}, \mathcal{F}_{t-1}]))),
\end{equation}
\begin{equation}
\bm{z_t} = \tanh(\texttt{Conv2d}_{W_{\mathcal{\bm{Z}}}}(\texttt{Concat}([\bm{\hat{{H_t^{g}}}}, \mathcal{F}_{t-1} \circ \bm{r_t}]))),
\end{equation}
where \( \bm{u_t} \) is the Update Gate, \( \bm{r_t} \) is the Reset Gate, \( {W_{\mathcal{\bm{U}}}}, {W_{\mathcal{\bm{R}}}},{W_{\mathcal{\bm{Z}}}} \) represent convolution kernel, and \( 
\bm{z_t} \) is the reset feature of the current moment. In this equation, \( \bm{\hat{{H_t^{g}}}} \) will be processed by a stride-2 convolution into a shape of : \(\bm{\hat{{H_t^{g}}}} \in \mathbb{R}^{T \times \frac{C}{4} \times \frac{H}{2} \times \frac{W}{2}}  \).

Then, the gate mechanism updates \( \mathcal{F'}_{t-1} \) as follows:
\begin{equation}
\mathcal{F'}_{t-1} = \mathcal{F}_{t-1} \circ (1 - \bm{u_t}) + \bm{z_t} \circ \bm{u_t}.
\end{equation}

Unlike the Global Focus Module, we only use adaptive average pooling to extract dynamic features from the hidden state \( \bm{\hat{{H_t^{g}}}} \):
\begin{equation}
\bm{G_{T}} = \mathcal{AAP}(\bm{\hat{{H_t^{g}}}}).
\end{equation}
Based on this dynamic feature, we calculate the dynamic balance factor \( \alpha_d \) and dynamic decay factor \( \beta_d \) through a lightweight linear projection:
\begin{equation}
\bm{\alpha_d}, \bm{\beta_d} = \sigma(\texttt{Linear}(\bm{G_{T}})).
\end{equation}

These two factors are designed to model the intrinsic physical processes of non-rigid motion. Specifically, \( \bm{\alpha} \) reflects the motion ``growth tendency", accounting for momentum preservation and internal deformation, while \( \bm{\beta} \) captures the ``dissipative decay", simulating energy loss over time due to friction or structural damping. Instead of assigning them as fixed hyperparameters, we adopt a data-driven approach that combines dynamic estimates \( \bm{\alpha_d}, \bm{\beta_d} \) with learnable priors \( \bm{\alpha'}, \bm{\beta'} \), using an element-wise arithmetic mean to ensure robustness:
\begin{equation}
\bm{\alpha} = \mathscr{T}(\bm{\alpha_d}, \bm{\alpha'}), \quad
\bm{\beta} = \mathscr{T}(\bm{\beta_d}, \bm{\beta'}).
\end{equation}
where \( \bm{\alpha'} \) and \( \bm{\beta'} \) are the initial parameters, respectively. \( \mathscr{T}(\cdot) \) is the transfer function. In this paper, the transfer function \( \mathscr{T}(\cdot) \) fuses the two inputs by computing their element-wise arithmetic mean. This fusion enables adaptive learning while preserving stable physical interpretability across different input distributions.

To further reflect the temporal attenuation of motion influence, we introduce a \textbf{Time Delay} mechanism:
\begin{equation}
\bm{D_{e}} = \exp(-\bm{\beta} \cdot \texttt{TimeSteps}), \label{eq:fifteen}
\end{equation}
where \texttt{TimeSteps} is a monotonically increasing time index tensor, \(\texttt{TimeSteps} \in \mathbb{R}^{T \times 1 \times 1 \times 1}\). The exponential decay formulation faithfully reflects physical systems where motion gradually weakens with time, such as damping oscillation or deformation relaxation. This allows the model to automatically prioritize recent motion cues and suppress the influence of outdated dynamics.

Object deformation can be summarized as the simultaneous action of growth and decay, so we adjust the motion trend based on the growth and decay processes:
\begin{equation}
\mathcal{D}_{t-1}^\prime = \bm{D_{e}} \cdot \mathcal{D}_{t-1} + \alpha \cdot \alpha_d \cdot \mathcal{D}_{t-1}.
\end{equation}
According to the physical scale, the motion trend is accumulated by the transient variations:
\begin{equation}
\mathcal{D}_t^M = 0.5 \mathcal{D}_{t-1}^\prime + 0.5 \mathcal{F}_{t-1},
\end{equation}
\begin{equation}
\mathcal{F}_t = \mathcal{D}_t^M + \mathcal{F}_{t-1}.
\end{equation}

Finally, the transient variations is applied to the input features:
\begin{equation}
\bm{X_t^{l}} = \texttt{DeformableConv}(\bm{\hat{{H_t^{g}}}}, \mathcal{F}_t).
\end{equation}

We use \texttt{DeformableConv} \cite{b25} to better adapt the motion variations. Compared with MotionGRU, MGM offers two key advantages. First, it explicitly models non-rigid deformations, enabling the capture of growth, dissipation, and morphological changes of targets during motion. Second, its factors have clear physical interpretations; combined with data-driven dynamic estimation and learnable priors, this allows the model to remain robust under varying input distributions and to effectively balance near-term and long-term information in video prediction over extended time horizons, thereby substantially improving its ability to capture complex dynamic targets and enhancing generalization performance. 

\begin{figure}[htbp]
\centerline{\includegraphics[width=10cm]{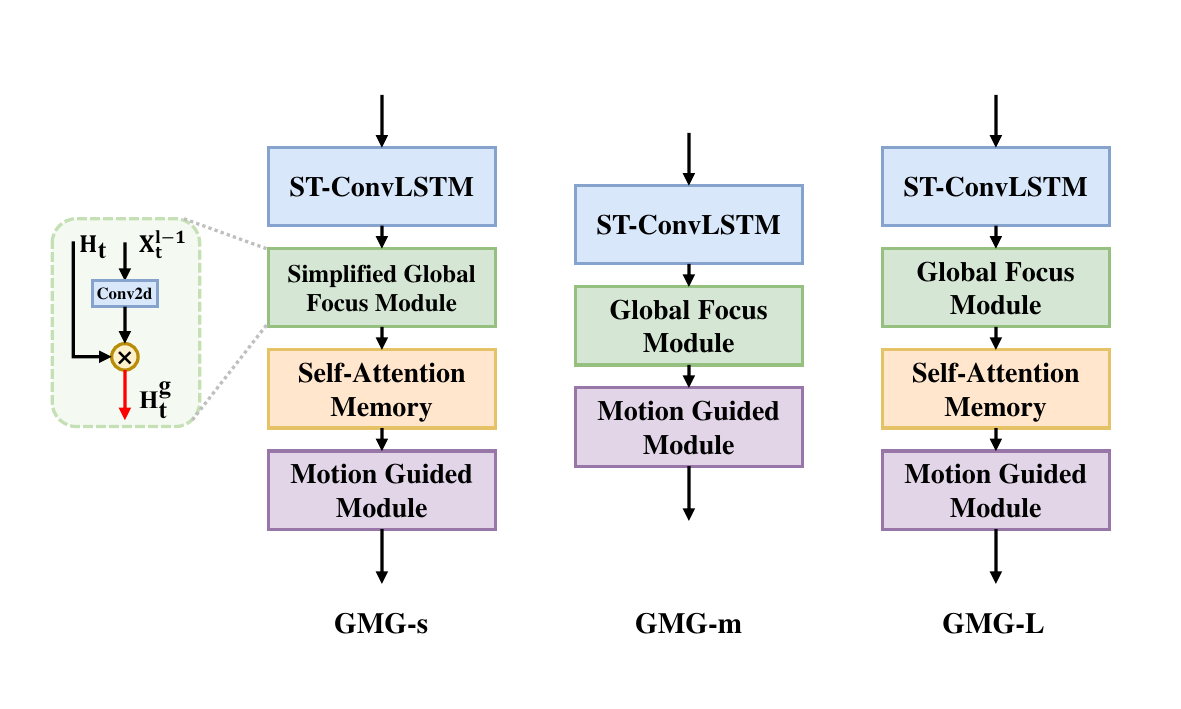}}
\caption{Diagram of GMG and Its Variants' Architecture.}
\label{fig:three}
\end{figure}

In addition, we propose two GMG variants to adapt to different tasks (GMG-s, GMG-m, refer to Fig. \ref{fig:three}). To accommodate large-sample datasets, a simplified GFM is provided to reduce computational complexity:

\begin{equation}
\bm{G_{F}} = \texttt{Conv2d}(\bm{X_t}),
\end{equation}
\begin{equation}
\bm{H_t^g} = \bm{H_t^g} \circ \bm{G_{F}}.
\end{equation}

The model that employs GFM-simpler is referred to as \textbf{GMG-s}. Additionally, to accommodate small-scale feature data, we omit the SAM module while retaining the GFM and other components, and this model is referred to as \textbf{GMG-m}. The base model introduced in Fig. \ref{fig:two} is named \textbf{GMG-L}.

To evaluate the efficiency of key modules, we compare their computational complexity and parameter sizes (see Table~\ref{tab:module_complexity}). The GFM module has a time complexity of $\mathcal{O}(BCHW)$ and a low memory footprint of $\mathcal{O}(BC)$, offering effective global modeling with high efficiency. In contrast, the MGM module is even more lightweight, with minimal computational and memory cost, making it well-suited for capturing fine-grained local motion.

In our model variants, \textbf{GMG-s} adopts a simplified version of GFM, while \textbf{GMG-m} removes the computationally expensive SAM module but retains GFM and MGM, achieving a good balance between performance and efficiency.

\begin{table}[ht]
\centering
\caption{Module complexity and size comparison.}
\small 
\setlength{\tabcolsep}{5pt} 
\begin{tabular}{lccccc}
\toprule
\textbf{Module} & \textbf{Time} & \textbf{Space} & \textbf{Params} & \textbf{FLOPs} \\
\midrule
SAM & $\mathcal{O}(BH^2W^2)$ & $\mathcal{O}(BH^2W^2)$ & 1.08M & 7.17G \\
GFM & $\mathcal{O}(BCHW)$ & $\mathcal{O}(BC)$ & 0.53M & 2.25G \\
MGM & $\mathcal{O}(BCHW)$ & $\mathcal{O}(BCHW)$ & 0.10M & 0.13G \\
\bottomrule
\end{tabular}
\label{tab:module_complexity}
\end{table}

\subsection{Overall structure for GMG}

The overall structure of the GMG model can be summarized as follows:
\begin{multline}
H_t,C_t,M_t = \\
\texttt{ST-ConvLSTM}(X_t^{l-1},H_{t-1},C_{t-1},M_{t-1}) 
\end{multline}
\begin{equation}
H_t^g = \texttt{GFM}(X_t^{l-1}, H_t)
\end{equation}
\begin{equation}
{\hat{H}}_t^g, {\hat{M}}_t = \texttt{SAM}(H_t^g, M_t)
\end{equation}
\begin{equation}
X_t^l, \mathcal{F}_t, \mathcal{D}_t^M = \texttt{MGM}({\hat{H}}_t^g, \mathcal{F}_{t-1}, \mathcal{D}_{t-1})
\end{equation}

Here, \(\texttt{GFM} \), \(\texttt{SAM} \), and \(\texttt{MGM} \) correspond to the Global Focus Module, Self-Attention Memory, and Motion Guided Module, respectively. Additionally, \(\texttt{ST-ConvLSTM} \) can be summarized as:

\begin{equation}
f_t = \sigma\left(W_{xf} * X_t + W_{hf} * H_{t-1} + b_f \right)
\end{equation}
\begin{equation}
i_t = \sigma\left(W_{xi} * X_t + W_{hi} * H_{t-1} + b_i \right)
\end{equation}
\begin{equation}
{\hat{C}}_t = \tanh\left(W_{xc} * X_t + W_{hc} * H_{t-1} + b_c \right)
\end{equation}
\begin{equation}
C_t = f_t \circ C_{t-1} + i_t \circ {\hat{C}}_t
\end{equation}
\begin{equation}
f_t' = \sigma\left(W_{xf'} * X_t + W_{hf'} * M_{t-1} + b_f' \right)
\end{equation}
\begin{equation}
i_t' = \sigma\left(W_{xi'} * X_t + W_{hi'} * {{M}}_{t-1} + b_i' \right)
\end{equation}
\begin{equation}
{\hat{C}}_t' = \tanh\left(W_{xc'} * X_t + W_{hc'} * M_{t-1} + b_c' \right)
\end{equation}
\begin{equation}
M_t = f_t' \circ {{M}}_{t-1} + i_t' \circ {\hat{C}}_t'
\end{equation}
\begin{multline}
o_t = \sigma\Big(W_{xo} * X_t + W_{ho} * {\hat{H}}_{t-1} + \\
W_{co} * C_t + W_{mo} * M_t + b_o \Big)
\end{multline}
\begin{equation}
H_t = o_t \circ \tanh\left(W_{1 \times 1} * \left[C_t, M_t \right]\right)
\end{equation}

where, * and \( \circ \) denote the convolution operator and the Hadamard product.

\section{Experiments}
\subsection{Datasets and Metrics}\label{AA}
In this paper, we conduct experiments on six datasets: \textbf{CIKM2017}\cite{b26}, \textbf{Shanghai2020}\cite{b27}, \textbf{Taxibj}\cite{b28}, \textbf{WeatherBench}\cite{b29}, \textbf{Moving MNIST}\cite{b31} and \textbf{Typhoon}\cite{b35}.

\textbf{CIKM2017}: This dataset, provided by the Shenzhen Meteorological Bureau, contains desensitized radar data with radar images at four altitudes. Each image has a resolution of \(101 \times 101\) square kilometers. The data is recorded at 6-minute intervals, covering 15 time steps (1.5 hours). In this study, we select 10,000 rainfall samples at the 0.5 km altitude. For each sample, the first 5 frames are used as input, and the subsequent 10 frames are predicted. The image resolution is reshaped to \(64 \times 64\). The dataset is divided into training and validation sets with a 9:1 ratio.

\textbf{Shanghai2020}: This dataset, published by the Shanghai Central Meteorological Observatory (SCMO) in 2020, records historical precipitation events in the Yangtze River Delta region over several years. Each sample consists of 20 consecutive radar echo frames spanning 3 hours, with the first 10 frames recorded at 6-minute intervals (as input) and the last 10 frames at 12-minute intervals (as prediction). A total of 9,000 samples are used for training and 1,000 for validation. The images are reshaped to \(64 \times 64\).

\textbf{Taxibj}: This dataset records traffic flow trajectories in Beijing, showing the inflow and outflow in each pixel of the regions using two channels. The training set contains 19,627 sample clips, and the test set contains 1,334 clips. Both training and test data have dimensions of \(2 \times 32 \times 32\). The model learns to predict the next 4 frames based on the previous 4 frames.

\textbf{WeatherBench}: For this study, we select global temperature data (T2m) from WeatherBench, which includes 52,559 training samples and 17,495 testing samples. The image size is \(32 \times 64\), and both input and prediction sequences consist of 12 frames with a 30-minute time interval.

\textbf{Moving MNIST}: This is one of the most popular datasets in the field of video prediction, consisting of two moving handwritten digits randomly sampled from the static MNIST dataset. The image resolution is \(64 \times 64\), with each digit sized at \(28 \times 28\). Each sequence contains 20 consecutive frames, where the first 10 frames serve as input and the last 10 frames as output. Both the training and test sets comprise 10,000 sequences.

\textbf{Typhoon}: The Digital Typhoon Dataset is a satellite image dataset designed for machine learning research on tropical cyclones. We randomly selected 10,140 samples from the Western Pacific (WP) and Around Australia (AU) regions as the training set, and 1,000 samples as the test set. Each sample consists of four input images and four output images, where each image has a spatial resolution of \(128 \times 128\) pixels and corresponds to a 1-hour interval in time. Compared with other datasets used in this study, this dataset further evaluates the model’s ability to predict fine-grained details.

\textbf{Implementation Details}: To ensure stable training and smooth predictions, the GMG model is optimized using \(L_2\) loss function, which penalizes larger errors more heavily and corresponds to the assumption of Gaussian-distributed noise. The specific model parameters, hyperparameters (Such as Batch size, Learning rate, Attention hidden layers and Training epochs), as well as the training hardware for each dataset, are detailed in Table \ref{tab:one}.

\textbf{Metrics}: We use common metrics to evaluate the predictive performance of the model, including MSE, MAE, RMSE, PSNR and SSIM. 

Specifically, for rainfall data, we also include the Critical Success Index (CSI) to assess forecast accuracy:

\begin{equation}
\text{CSI} = \frac{TP}{TP + FP + FN}
\end{equation}

where \( TP \) represents true positives (correctly predicted positive events), \( FP \) denotes false positives (events predicted as positive but actually negative), and \( FN \) stands for false negatives (events that were not predicted but are actually positive). A higher CSI value indicates better model performance, with a maximum possible score of 1. CSI is widely used in applications such as weather forecasting, event prediction, and video anomaly detection.

\begin{table*}[ht]
\centering
\small
\caption{Experimental settings for different datasets.}
\label{tab:one}
\resizebox{1.0\textwidth}{!}{%
\begin{tabular}{@{}l|cccccccc@{}}
\toprule
\textbf{Dataset}    & \textbf{Patch / Batch size} & \textbf{Resolution} & \textbf{Model Type} &\textbf{Task}           & \textbf{Epochs} & \textbf{Learning Rate} & \textbf{Att Hidden} & \textbf{GPU}        \\ \midrule
CIKM2017            & 4/16                          & (1×64×64)     &    GMG-L      & \(5 \rightarrow 10\)    & 100             & 0.0003                 & 32                  & 1 * RTX4060Ti         \\ 
Shanghai2020        & 4/16                          & (1×64×64)     &    GMG-L      & \(10 \rightarrow 10\)   & 200             & 0.0003                 & 32                  & 1 * RTX4060Ti         \\ 
Taxibj              & 2/16                          & (2×32×32)    &    GMG-m       & \(4 \rightarrow 4\)     & 200             & 0.0003                 & 64                  & 1 * RTX4060Ti         \\ 
WeatherBench        & 2/16                          & (1×32×64)     &    GMG-s      & \(12 \rightarrow 12\)   & 100             & 0.0003                 & 64                  & 4 * TESLA V100              \\ 
Moving MNIST        & 4/16                          & (1×64×64)     &    GMG-L      & \(10 \rightarrow 10\)   & 600/2000             & 0.0003                 & 32                  & 1 * TESLA V100              \\
Typhoon        & 4/16                          & (1×128×128)     &    GMG-L      & \(4 \rightarrow 4\)   & 100             & 0.0003                 & 64                  & 1 * TESLA V100              \\
 \bottomrule
\end{tabular}%
}
\end{table*}

\subsection{Quantitative Evaluation}

To analyze the performance of GMG compared to other SOTA models, we summarize the experimental results. Tables \ref{tab:two}, \ref{tab:three}, \ref{tab:four}, and \ref{tab:five} present these quantitative results. Numbers highlighted in \textcolor{red}{\textbf{Red Bold}} indicate the best performance for the corresponding metric, while numbers marked with a \textcolor{blue}{\underline{Blue Underlined}} represent the second-best performance. In this study, we select deterministic models for comparison, including CNN, RNN, CNN-RNN, and ViT-based models. The selected models also include mainstream SOTA models such as SimVP and its variants, SwinLSTM, and WaST.

Table \ref{tab:two} presents the performance of baseline models and GMG on the CIKM2017 and Shanghai2020 datasets. GMG achieves leading performance across all metrics. Compared to the best-performing baseline models, GMG reduces the MSE by 8\% on CIKM2017 (Compare with MIM) and 4.1\% on Shanghai2020 (Compare with PredRNN-V2). Additionally, SSIM improves by 0.0107 on CIKM2017 and 0.0012 on Shanghai2020. Notably, RNN-based models (e.g., MIM, MotionRNN, PredRNN-V2) tend to perform better in these two rainfall datasets. This may be attributed to the superior temporal modeling capability, spatial information capture, and effective long-term dependency modeling of recurrent neural networks.

Table \ref{tab:three} compares the performance of different models on the TaxiBJ dataset. Compared to the current best model, WaST, GMG achieves a new SOTA in MAE, SSIM, and PSNR. Compared to the baseline model MotionRNN, GMG reduces MSE by 20.63\% and lowers MAE by approximately 7.9\%.

Tables \ref{tab:four} and \ref{tab:five} present the experimental results on the WeatherBench and Moving MNIST datasets, respectively, where GMG achieves comprehensive superiority across all evaluation metrics. With minor modifications, GMG, as an RNN-based model, outperforms the currently popular CNN and ViT models across different tasks. This provides a new perspective on utilizing recurrent neural networks for complex video prediction tasks in the future.

Specifically, on the Moving MNIST dataset, we trained the GMG model for 2000 epochs, and the results show a significant performance improvement. Compared to GMG trained for a standard (shorter) number of epochs, the longer-trained GMG achieved noticeable gains across all evaluation metrics. This further demonstrates the stability and generalization capability of the GMG architecture under extended training, providing strong support for further optimization of complex video prediction tasks.

Table \ref{tab:six} presents the quantitative results on the Typhoon dataset. GMG achieves the best performance across all evaluation metrics. Compared to the MotionRNN, GMG reduces the MSE by 1.2\% and reduced RMSE from 12.67 to 12.60. Furthermore, GMG surpasses WaST in both MSE and MAE. Notably, Typhoon is a higher-resolution dataset with complex and chaotic motion patterns, yet GMG maintains its leading performance, demonstrating strong generalization and robustness in challenging real-world scenarios.

\begin{table*}[ht]
\centering
\scriptsize
\caption{Quantitative results on CIKM2017 and Shanghai2020 datasets.}
\label{tab:two}
\resizebox{0.95\textwidth}{!}{%
\begin{tabular}{@{}c|l|ccccccc@{}}
\toprule
\textbf{Dataset} & \textbf{Model}            & \textbf{Params} & \textbf{MSE (↓)} & \textbf{MAE (↓)} & \textbf{SSIM (↑)} & \textbf{CSI30 (↑)} & \textbf{CSI40 (↑)} & \textbf{CSI50 (↑)} \\ \midrule
\textbf{\multirow{14}{*}{\rotatebox{90}{CIKM2017}}} & ConvLSTM (NeurIPS'2015)\cite{b23}          & 15.1M  & 28.8162 & 161.242 & 0.7359    & 0.7757     & 0.6590     & 0.5367     \\
                           & PredRNN (NeurIPS'2017)\cite{b4}          & 23.8M  & 28.8126 & 159.404 & 0.7303    & 0.7803     & 0.6690     & 0.5471     \\
                           & PredRNN++ (ICML'2018)\cite{b5}         & 38.6M  & 28.0987 & 159.258 & 0.7362    & 0.7775     & 0.6684     & 0.5533     \\
                           & MIM (CVPR'2019)\cite{b6}              & 38.04M & \textcolor{blue}{\underline{27.2072}} & \textcolor{blue}{\underline{154.269}} & 0.7388    & 0.7828     & 0.6725     & \textcolor{blue}{\underline{0.5587}}     \\
                           & PyDNet (CVPR'2020)\cite{b15}          & 3.1M   & 28.9757 & 161.143 & 0.7393    & 0.7809     & 0.6654     & 0.5489     \\
                           & MotionRNN (CVPR'2021)\cite{b7}       & 26.9M  & 27.2091 & 155.827 & \textcolor{blue}{\underline{0.7406}}    & \textcolor{blue}{\underline{0.7867}}     & \textcolor{blue}{\underline{0.6762}}    & 0.5510     \\
                           & MAU (NeurIPS’2021)\cite{b21}              & 4.48M  & 30.9094 & 167.361 & 0.7234    & 0.7731     & 0.6584     & 0.5459     \\
                           & PredRNN-V2 (TPAMI'2022)\cite{b8}       & 23.9M  & 28.0085 & 161.508 & 0.7374    & 0.7857     & 0.6640     & 0.5362     \\
                           & SimVP-gSTA (CVPR'2022)\cite{b11}        & 4.82M  & 31.3121 & 166.261 & 0.7195    & 0.7731     & 0.6594     & 0.5416     \\
                           & SimVP-ViT (NeurIPS'2023)\cite{b17}       & 39.6M  & 27.8803 & 157.478 & 0.7358    & 0.7835     & 0.6669     & 0.5483     \\
                           & Swin-LSTM (ICCV'2023)\cite{b16}         & 20.19M     & 27.9612 & 158.899 & 0.7405 & 0.7826 & 0.6652 & 0.5532    \\
                           & TAU (CVPR'2023)\cite{b12}             & 38.4M  & 30.5827 & 161.948 & 0.7277    & 0.7779     & 0.6601     & 0.5370     \\
                           & WaST (AAAI'2024)\cite{b30}            & 28.18M & 30.3074& 165.804 & 0.7309 & 0.7773 & 0.6534 & 0.5174 \\
                           & \textbf{GMG (Ours)}   & 31.44M & \textcolor{red}{\textbf{25.0215}} & \textcolor{red}{\textbf{149.493}} & \textcolor{red}{\textbf{0.7513}}    & \textcolor{red}{\textbf{0.7885}}     & \textcolor{red}{\textbf{0.6812}}     & \textcolor{red}{\textbf{0.5682}}     \\ \midrule
\textbf{\multirow{14}{*}{\rotatebox{90}{Shanghai2020}}} & ConvLSTM (NeurIPS'2015)\cite{b23}         & 15.1M  & 5.0219 & 39.236 & 0.9162    & 0.4321     & 0.4046     & 0.3584     \\
                               & PredRNN (NeurIPS'2017)\cite{b4}          & 23.8M  & 4.3347 & \textcolor{red}{\textbf{34.297}} & \textcolor{blue}{\underline{0.9288}}    & \textcolor{blue}{\underline{0.4707}}     & \textcolor{blue}{\underline{0.4451}} & \textcolor{blue}{\underline{0.3992}}     \\
                               & PredRNN++ (ICML'2018)\cite{b5}           & 38.6M  & 4.7445      & 39.416      & 0.9190         & 0.4419          & 0.4100          & 0.3605          \\
                               & MIM (CVPR'2019)\cite{b6}                 & 38.04M & 6.3924 & 44.986 & 0.8997    & 0.3910     & 0.3595     & 0.3131     \\
                               & PyDNet (CVPR'2020)\cite{b15}              & 3.1M   & 7.6126      & 50.155      & 0.8846         & 0.3508          & 0.3124          & 0.2663          \\
                               & MotionRNN (CVPR'2021)\cite{b7}           & 26.9M  & 4.5867 & 37.871 & 0.9221    & 0.4457     & 0.4183     & 0.3708     \\
                               & MAU (NeurIPS’2021)\cite{b21}              & 4.48M  & 7.3441 & 50.766 & 0.8853    & 0.3485     & 0.3098     & 0.2599     \\
                               & PredRNN-V2 (TPAMI'2022)\cite{b8}         & 23.9M  & \textcolor{blue}{\underline{4.2050}} & \textcolor{blue}{\underline{35.015}} & 0.9270    & 0.4543     & 0.4272     & 0.3815     \\
                               & SimVP-gSTA (CVPR'2022)\cite{b11}          & 4.82M  & 8.0889 & 45.846 & 0.8908    & 0.3649     & 0.3406     & 0.2991     \\
                               & SimVP-ViT (NeurIPS'2023)\cite{b17}        & 39.6M  & 9.9074 & 52.900 & 0.8718    & 0.3272     & 0.3053     & 0.2667     \\
                               & Swin-LSTM (ICCV'2023)\cite{b16}         & 20.19M     & 6.7183 & 46.025 & 0.8957 & 0.3714 & 0.3371 & 0.2902    \\
                               & TAU (CVPR'2023)\cite{b12}                 & 38.4M  & 8.2874 & 47.315 & 0.8886    & 0.3630     & 0.3421     & 0.3000     \\
                               & WaST (AAAI'2024)\cite{b30}            & 28.18M & 6.1937 & 42.387 & 0.9063 & 0.3908 &  0.3545 & 0.3042\\
                               & \textbf{GMG (Ours)}           & 31.44M & \textcolor{red}{\textbf{4.0308}} & 35.771 & \textcolor{red}{\textbf{0.9300}}    & \textcolor{red}{\textbf{0.4741}}     & \textcolor{red}{\textbf{0.4487}}     & \textcolor{red}{\textbf{0.4002}}     \\ \bottomrule
\end{tabular}%
}
\end{table*}

\begin{table}[ht]
\centering
\small
\caption{Quantitative results in Taxibj datasets.}
\label{tab:three}
\resizebox{0.5\textwidth}{!}{%
\begin{tabular}{@{}l|cccc@{}}
\toprule
\textbf{Model}             & \textbf{MSE × 100 (↓)} & \textbf{MAE (↓)} & \textbf{SSIM (↑)}  & \textbf{PSNR (↑)} \\ \midrule
ConvLSTM\cite{b23}         & 40.0678 & 16.1892 & 0.9819 & 38.888   \\
PredRNN\cite{b4}           & 35.0348 & 15.1302 & 0.9844 & 39.591  \\
PredRNN++\cite{b5}            & 41.8227 & 15.9766 & 0.9824 & 39.135   \\
PyDNet\cite{b15}               & 40.1700 & 16.4790 & 0.9808 & 38.939   \\
MotionRNN\cite{b7}            & 37.6517 & 16.0009 & 0.9825 & 39.001   \\
MAU\cite{b21}               & 40.7206 & 15.6620 & 0.9822 & 39.353   \\
PredRNN-V2\cite{b8}          & 45.2737 & 16.6105 & 0.9807 & 38.713   \\
SimVP-gSTA\cite{b11}            & 36.7385 & 15.3530 & 0.9832 & 39.509   \\
Swin-LSTM\cite{b16}            & 35.9456 & 15.2276 & 0.9832 & 39.645  \\
TAU\cite{b12}                   & 35.1037 & 15.1745 & 0.9838 & 39.593   \\  
WasT\cite{b30}                  & \textbf{\textcolor{red}{29.7753}}  & \textcolor{blue}{\underline{14.7945}} & \textcolor{blue}{\underline{0.9846}} & \textcolor{blue}{\underline{39.777}} \\
\textbf{GMG (Ours)}           & \textcolor{blue}{\underline{29.8812}} & \textbf{\textcolor{red}{14.7277}} & \textbf{\textcolor{red}{0.9850}} & \textbf{\textcolor{red}{39.831}} \\ \bottomrule
\end{tabular}%
}
\end{table}

\begin{table}[ht]
\centering
\small
\caption{Quantitative results in WeatherBench datasets.}
\label{tab:four}
\resizebox{0.45\textwidth}{!}{%
\begin{tabular}{@{}l|ccc@{}}
\toprule
\textbf{Model}             & \textbf{MSE (↓)} & \textbf{MAE (↓)} & \textbf{RMSE (↓)} \\ \midrule
ConvLSTM\cite{b23}           & 1.9703           & 0.8472           & 1.4036           \\
PredRNN\cite{b4}                & 1.2881              & 0.7035              &1.1349             \\
MIM\cite{b6}                       & 1.8480              & 0.8611              & 1.3594              \\
MotionRNN\cite{b7}                & \textcolor{blue}{\underline{1.2607}}              & 0.6813              & \textcolor{blue}{\underline{1.1228}}              \\
MAU\cite{b21}                   & 1.4381              & 0.7548              & 1.1992              \\
PredRNN-V2\cite{b8}            & 1.8430           & 0.9029           & 1.3575           \\
SimVP-gSTA\cite{b11}            & 1.5934           & 0.7654           & 1.2623           \\
TAU\cite{b12}                     & 1.4986  & 0.7375  & 1.2241  \\
WaST\cite{b30}                 & 1.3387              & \textcolor{blue}{\underline{0.6808}}              & 1.1570              \\
\textbf{GMG (Ours)}                      & \textbf{\textcolor{red}{1.2341}}              & \textbf{\textcolor{red}{0.6780}}              & \textbf{\textcolor{red}{1.1109}}              \\  \bottomrule
\end{tabular}%
}
\end{table}

\begin{table}[ht]
\centering
\small
\caption{Quantitative results in Moving MNIST datasets. (GMG* denotes the results obtained after 2000 training epochs.)}
\label{tab:five}
\resizebox{0.5\textwidth}{!}{%
\begin{tabular}{@{}l|cccc@{}}
\toprule
\textbf{Model}                        & \textbf{MSE (↓)} & \textbf{MAE (↓)} & \textbf{SSIM (↑)} & \textbf{PSNR (↑)} \\ \midrule
SimVP-gSTA\cite{b11}                      & 22.5268           & 67.8671           & 0.9500           & 23.6783           \\
SimVP-ViT \cite{b17}                   & 26.4819           & 77.5663           & 0.9371           & 23.0279           \\
SimVP-VAN \cite{b17}                   & 20.5918           & 63.5674           & 0.9547           & 24.1267           \\
SimVP-Poolformer\cite{b17}           & 25.5146           & 74.6528           & 0.9429           & 23.1193           \\
Swin-LSTM\cite{b16}                  & 19.4554           & 61.2669           & 0.9571           & 24.3593           \\
TAU  \cite{b12}                        & 19.9112          & 62.1182          & 0.9562           & 24.3096           \\
WasT\cite{b30}                          &  22.0719 & 70.8779 & 0.9491 & 23.7451 \\
\textbf{GMG (Ours)}                               & \textcolor{blue}{\underline{19.0741}}           & \textcolor{blue}{\underline{60.7413}}           & \textcolor{blue}{\underline{0.9586}}           & \textcolor{blue}{\underline{24.4606}}           \\ 
\textbf{GMG* (Ours)}                               & \textbf{\textcolor{red}{14.2479}}           & \textbf{\textcolor{red}{48.1517}}           & \textbf{\textcolor{red}{0.9700}}           & \textbf{\textcolor{red}{26.3387}}           \\
\bottomrule
\end{tabular}%
}
\end{table}

\begin{table}[ht]
\centering
\small
\caption{Quantitative results in Typhoon datasets.}
\label{tab:six}
\resizebox{0.45\textwidth}{!}{%
\begin{tabular}{@{}l|ccc@{}}
\toprule
\textbf{Model} & \textbf{MSE (↓)} & \textbf{MAE/10 (↓)} & \textbf{RMSE (↓)} \\
\midrule
ConvLSTM\cite{b23}  & 224.389 & 125.000 & 14.9796 \\
PredRNN\cite{b4} & 216.077 & 122.643 & 14.6995 \\
PredRNN++\cite{b5} & 183.307 & 113.522 & 13.539 \\
MIM\cite{b6} & 222.208 & 124.240 & 14.9066 \\
MotionRNN\cite{b7} &  \textcolor{blue}{\underline{160.656}} & 107.228 &  \textcolor{blue}{\underline{12.6750}} \\
MAU\cite{b21}  & 166.581 & 109.660 & 12.9066 \\
PredRNN-V2\cite{b8} & 183.416 & 114.336 & 13.5431 \\
SimVP-gSTA\cite{b11} & 204.329 & 119.063 & 14.2943 \\
TAU\cite{b12} & 197.883 & 116.552 & 14.0671 \\
WaST\cite{b30} & 168.903 &  \textcolor{blue}{\underline{106.952}} & 12.9962 \\
\textbf{GMG (Ours)} & \textbf{\textcolor{red}{158.794}} & \textbf{\textcolor{red}{106.381}} & \textbf{\textcolor{red}{12.6013}} \\
\bottomrule
\end{tabular}%
}
\end{table}

\subsection{Qualitative Evaluation}

\begin{figure}[htbp]
\centerline{\includegraphics[width=9.0cm]{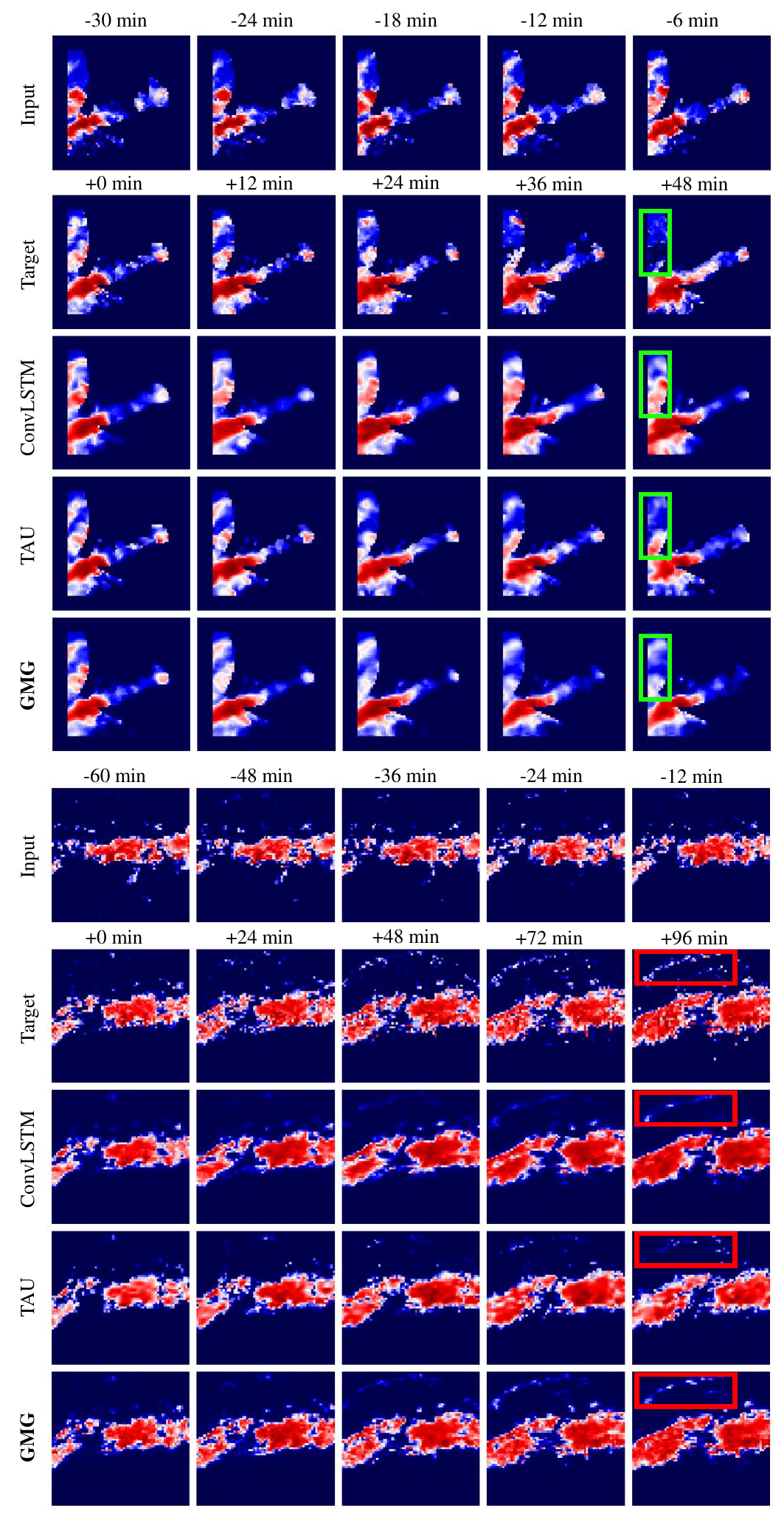}}
\caption{Visualizations on CIKM2017 (upper) and Shanghai2020 (lower). ``+min'' indicates the forecast lead time of the predicted images, while ``-min'' represents the time of the input images.
}
\label{fig:four}
\end{figure}

To further analyze the model's performance in rainfall prediction, we visualize the results (Fig. \ref{fig:four}). In the green-boxed region, both ConvLSTM and TAU exhibit varying degrees of false alarms, while GMG accurately predicts rainfall in this area. Additionally, in the red-boxed region, neither ConvLSTM nor TAU detects the rainfall, but GMG successfully captures the development of rainfall from absence to presence. This highlights the significant contributions of the Global Focus and Motion Guided modules in improving prediction performance.

\begin{figure}[htbp]
\centerline{\includegraphics[width=9.0cm]{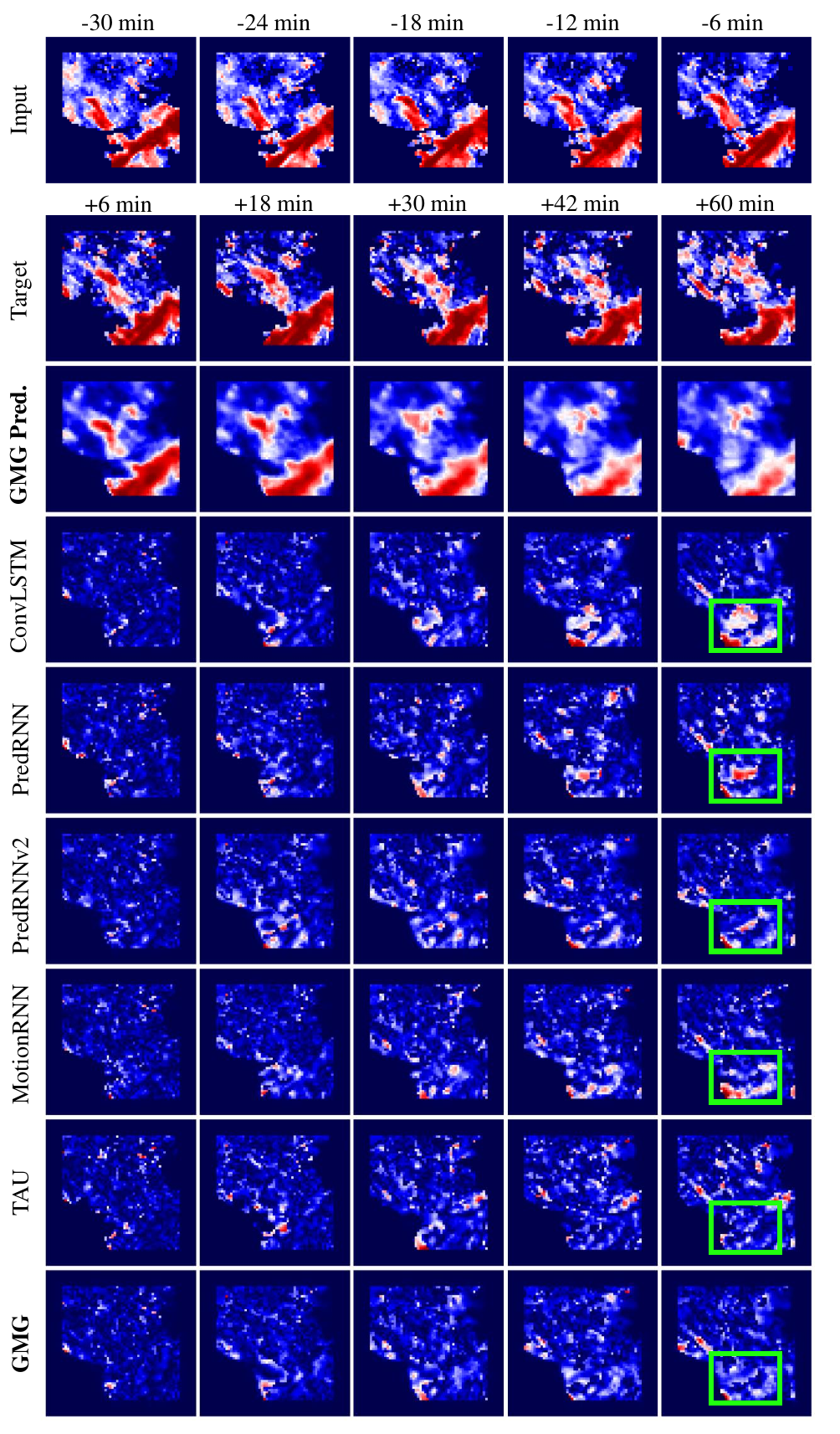}}
\caption{Error visualizations on CIKM2017 dataset, ``GMG Pred.'' indicate the real prediction results made by GMG.
}
\label{fig:b1}
\end{figure}

Fig. \ref{fig:b1} presents the error analysis for a typical case from the CIKM 2017 dataset. Compared to the Shanghai 2020 dataset, this dataset is more challenging to predict, yet GMG, through its ability to adaptively model the dynamic evolution of non-rigid meteorological patterns, is able to effectively correct predictions in regions with larger errors. Despite having limited input information, GMG is still able to make accurate predictions for rainfall areas, capturing the changes in precipitation processes and the movement of clouds, which are complex weather phenomena.

\begin{figure}[htbp]
\centerline{\includegraphics[width=9.0cm]{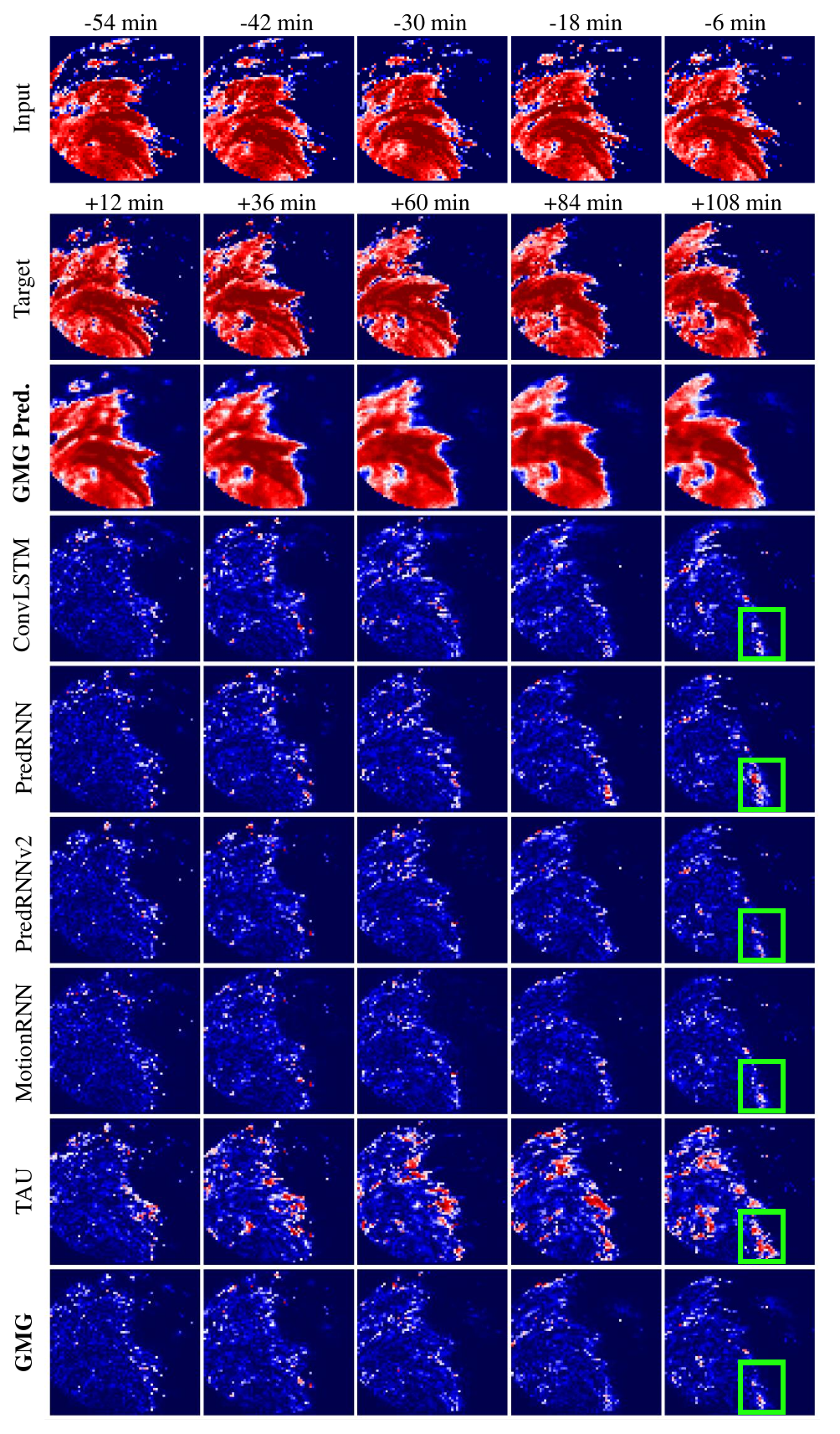}}
\caption{Error visualizations on Shanghai2020 dataset, ``GMG Pred.'' indicate the real prediction results made by GMG.
}
\label{fig:b2}
\end{figure}

Fig. \ref{fig:b2} presents the error analysis for a typical case from Shanghai2020. Through visual comparison with the \textcolor{green}{green} box, it is evident that the GMG model exhibits smaller errors in more complex regions compared to other models. The prediction results from GMG show clear details with smooth edge transitions. Compared to traditional CNN or RNN models, GMG demonstrates stronger capability in detail reconstruction in these critical areas, resulting in smaller errors.

\begin{figure*}[htbp]
\centerline{\includegraphics[width=18cm]{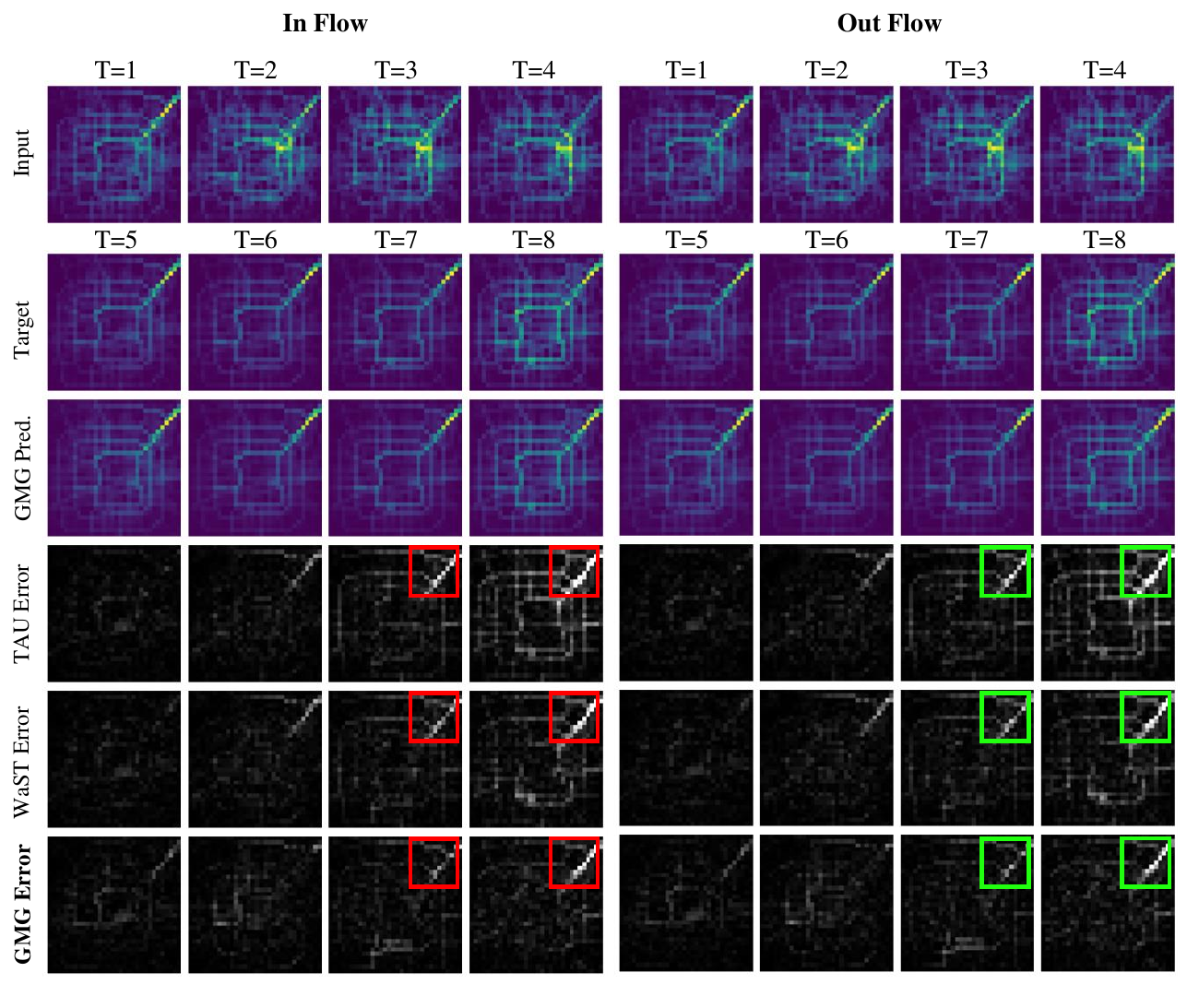}}
\caption{Visualizations on Taxibj, Error = $\left| \text{Prediction} - \text{Target} \right|$ , we amplify the error for better comparison.
}
\label{fig:five}
\end{figure*}

Fig. \ref{fig:five} presents the experimental results on Taxibj, comparing the outcomes of TAU and WaST. In the main regions where errors occur (marked by \textcolor{green}{green} and \textcolor{red}{red} boxes), GMG consistently maintains the lowest error. This further demonstrates the effectiveness of the Global Focus Module in accurately capturing long-range dependencies.

\begin{figure*}[htbp]
\centerline{\includegraphics[width=14cm]{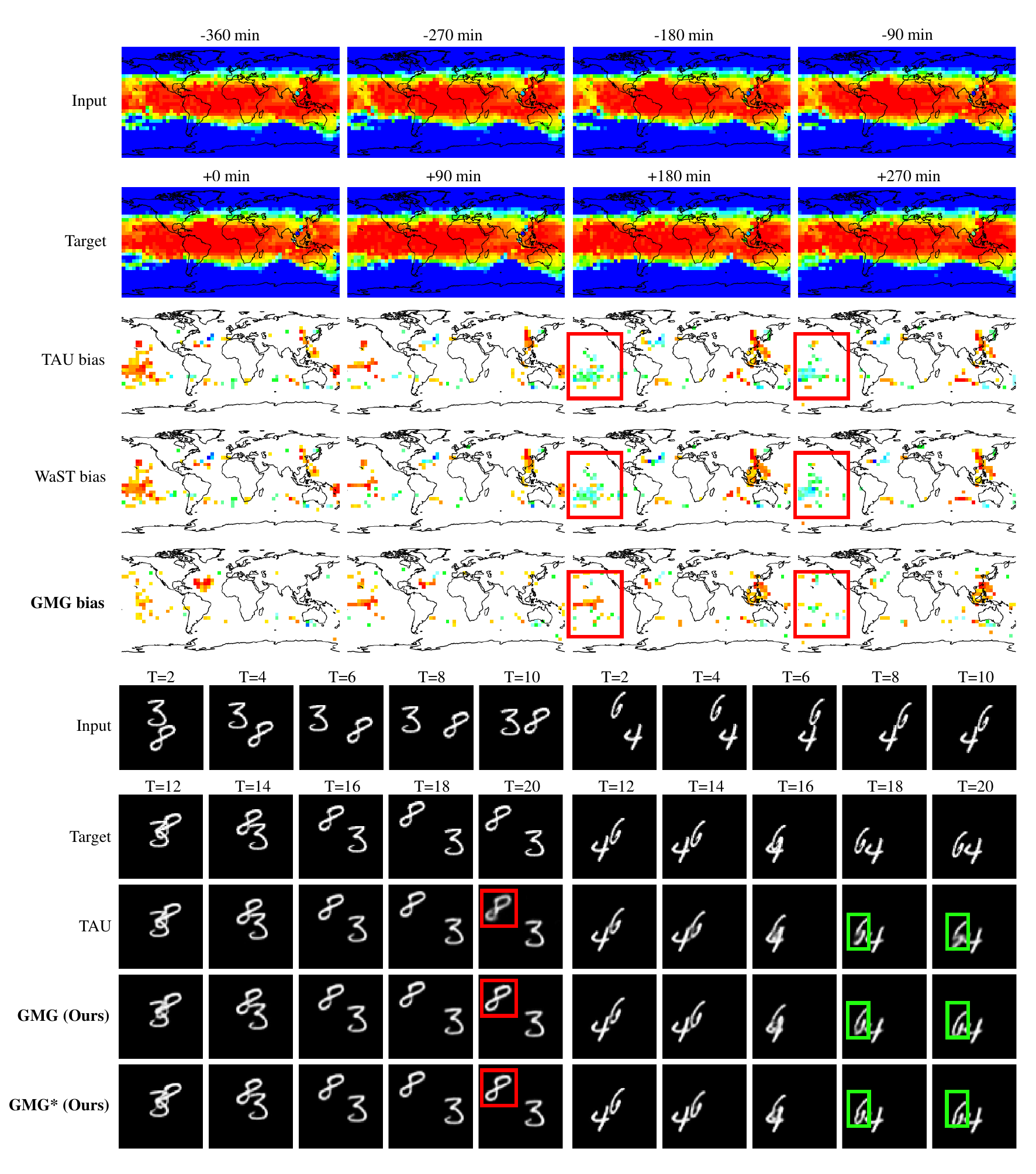}}
\caption{Visualizations on WeatherBench (upper) and Moving MNIST datasets (lower) , bias = \text{Prediction} - \text{Target}, which can reflect the extent to which the model tends to predict colder or warmer temperatures.
}
\label{fig:six}
\end{figure*}

Fig.~\ref{fig:six} presents the visualization results of GMG on the WeatherBench and Moving MNIST datasets. In the WeatherBench section, we display the ground truth temperature maps, GMG’s predictions, and the corresponding error maps in comparison with other models such as TAU and WasT. We specifically highlight the main error-prone regions over the equatorial Pacific area with red boxes. It is clearly observed that GMG achieves the smallest prediction errors in these sensitive regions, with error colors tending toward green (indicating lower error), while other models exhibit more significant deviations. This demonstrates that even though temperature variations in WeatherBench are relatively gradual, GMG leveraging the Global Focus Module and Motion Guided Module can still accurately capture subtle temperature changes, showcasing strong modeling capability.

In the Moving MNIST section, we compare the performance of GMG at different training stages (600 epochs vs. 2000 epochs, the latter denoted as GMG*) and analyze two representative examples: one involving the digit ``8'' and the other the digit ``6''.

In the case of digit ``8'', we observe that the TAU model generates blurry and structurally broken predictions in later time steps ($T=8\sim10$), especially failing to maintain the closed loop of the central ring (as highlighted by red boxes). In contrast, GMG trained with 600 epochs is already able to preserve the overall shape and loop closure of the digit ``8'', though some edge blur and detail loss remain. GMG* (trained for 2000 epochs), however, produces much sharper and more coherent predictions, with clear boundaries and complete structure—almost indistinguishable from the ground truth. This indicates that longer training significantly enhances the model’s ability to capture fine-grained details and improves prediction stability.

In the case of digit ``6'', TAU’s predictions suffer from deformation and displacement, particularly around the lower curved region of the digit. GMG (600 epochs) performs better, preserving the general outline of the digit ``6'', although the stroke appears slightly thin and lacks fullness. GMG* further improves this by accurately restoring the shape, maintaining consistent thickness and spatial continuity (as shown by green boxes), suggesting the model’s strong ability to track rigid object contours and generate more natural and stable predictions.

\begin{figure}[htbp]
\centerline{\includegraphics[width=9cm]{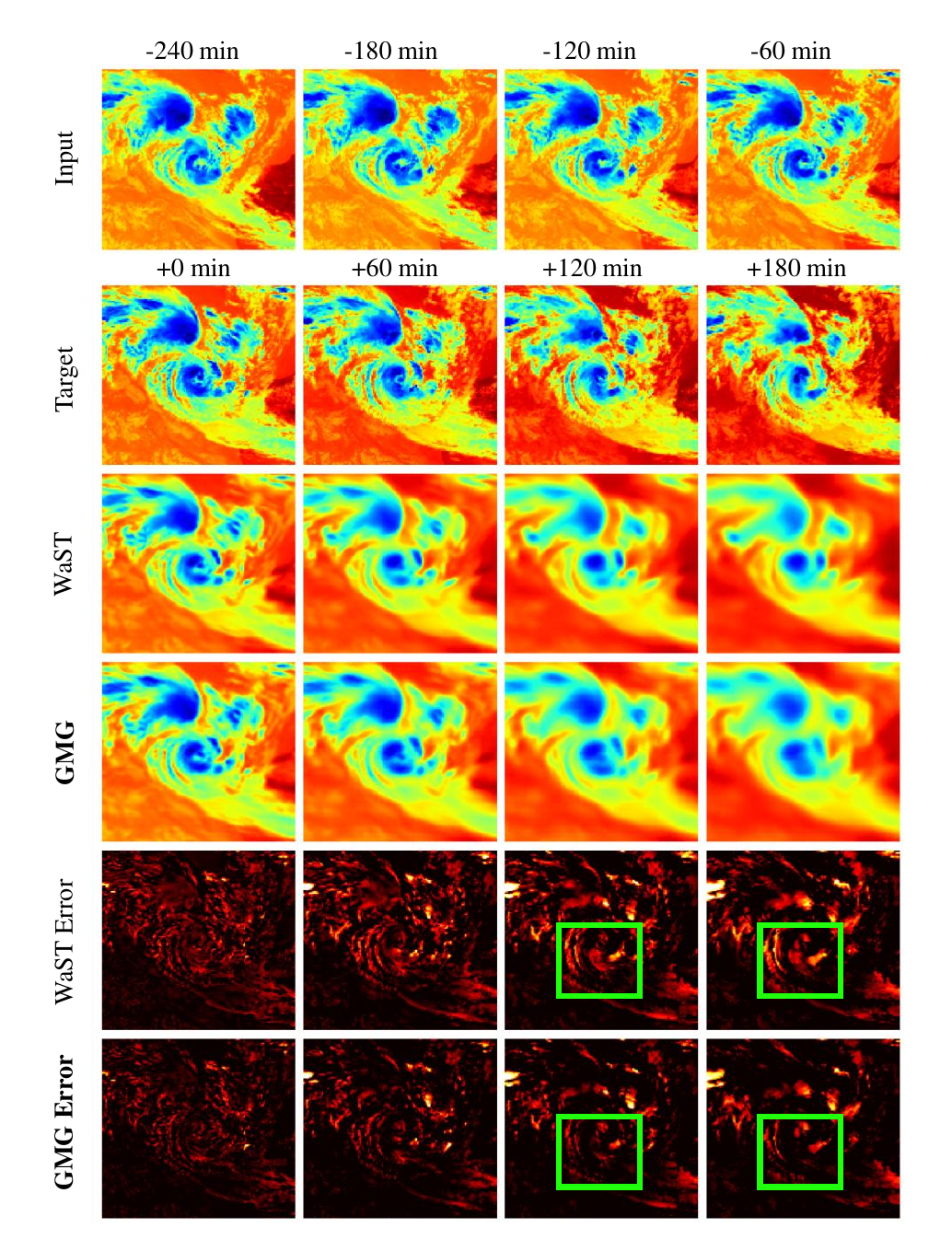}}
\caption{Visualizations on Typhoon, Error = $\left| \text{Prediction} - \text{Target} \right|$ , we amplify the error for better comparison.
}
\label{fig:b3}
\end{figure}

Fig. \ref{fig:b3} presents the visualization results for Typhoon. The visualization results show that GMG performs exceptionally well on datasets with high resolution and complex dynamic patterns, such as typhoons. Compared to WaST, GMG exhibits lower and more localized prediction errors in key regions like the typhoon eye and strong convective bands, preserving spatial details more effectively. Additionally, GMG produces more temporally coherent forecasts across consecutive time steps, demonstrating stronger sequence modeling capabilities, making it well-suited for fine-grained prediction of complex weather systems.

From the above visualization results, we observe that GMG can adapt to various types of video prediction tasks, regardless of whether the predicted subject is a rigid or non-rigid body. Moreover, it achieves superior performance in terms of prediction accuracy and object detail preservation. This demonstrates that GMG can serve as a general-purpose video prediction model, making it applicable to real-world complex video prediction tasks.

\subsection{Ablation Study}

To validate the contributions of GMG's individual components, we conduct ablation experiments on the CIKM2017 dataset, with results presented in Table \ref{tab:seven}. In addition to controlled experiments on key components—GFM, MGM, and SAM—we include comparisons with baseline models, PredRNN and MotionRNN. The results show that adding GFM, SAM, or MGM improves predictive performance over the baselines. Furthermore, compared to MotionGRU, model (A), which incorporates the MGM module to account for growth or dissipation processes, better captures rainfall data characteristics. By comparing Model (A), (B), and (C), MGM emerges as the most impactful component in enhancing baseline performance, followed by GFM, further validating the effectiveness of the proposed model design. Model (D) incorporates both the GFM and SAM modules, and even without using MGM, it achieves a significant performance improvement, indicating the positive roles of GFM and SAM in feature extraction and spatial modeling. Additionally, the lightweight version, GMG-s, achieves competitive results, offering a viable pathway for future lightweight modeling efforts. We also provide the parameter count and training time for each model, offering a reference for practical model deployment.

To compare the impact of each module on training and inference efficiency, we report the parameter count (Params), computational complexity (FLOPs), and inference speed (FPS, Frames Per Second) of each model (Table \ref{tab:seven}). The results show that although the GFM and MGM modules are introduced, the full model GMG-L only slightly increases in parameters and FLOPs compared to the baseline, while maintaining an inference speed of 49.11 FPS and a training complexity within an acceptable range. The lightweight version GMG-s achieves excellent accuracy with a fast inference speed of 52.21 FPS, indicating that the proposed modules significantly improve model performance with minimal impact on training and inference efficiency, demonstrating strong practicality and deployment value.

\begin{figure}[htbp]
\centerline{\includegraphics[width=9.0cm]{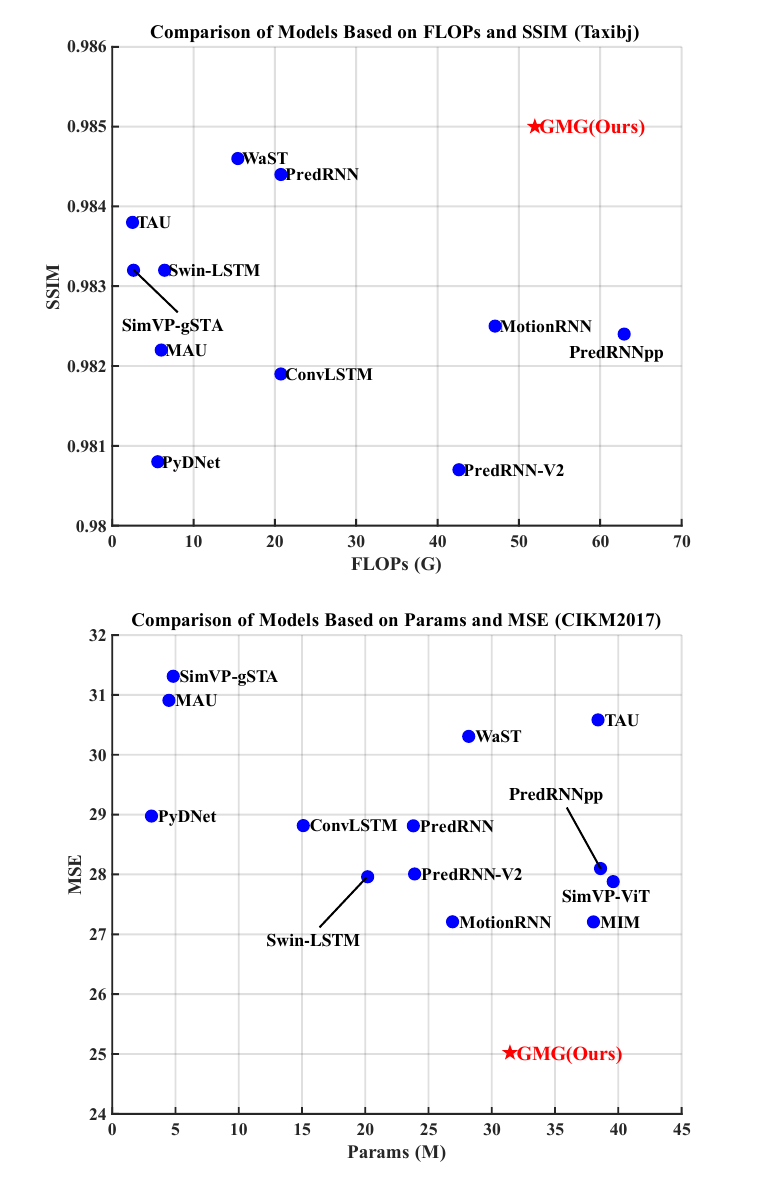}}
\caption{ The comparison for model performance and efficiency
}
\label{fig:seven}
\end{figure}

Additionally, we visualize the performance-parameter relationship of the models. Fig. \ref{fig:seven} presents a comprehensive comparison of all models' performance on CIKM2017 and Taxibj, showing that GMG effectively balances model parameters, computational cost, and performance, achieving the best overall performance.

To analyze how the number of GMG Cell layers affects model performance, we conducted experiments comparing configurations with 2 to 5 layers. As shown in Table \ref{tab:eight}, increasing the number of layers from 2 to 4 led to notable improvements in accuracy, with the 4-layer setup achieving the best results (MSE of 4.0308 and SSIM of 0.9300). However, further increasing to 5 layers resulted in a slight decline in prediction quality, while the computational cost increased significantly (FLOPs reached 0.196T and FPS dropped to 21.17). These results suggest that model depth plays a key role in balancing predictive accuracy and computational efficiency, and highlight the performance trade-offs involved in deeper architectures.

To further verify the effectiveness of the proposed Global Focus Module (GFM), we conduct ablation experiments on ConvRNN-based models using the CIKM2017 and TaxiBJ datasets, with results shown in Table \ref{tab:gfm_ablation} and Table \ref{tab:gfm_ablation2}. The experimental results demonstrate that introducing GFM consistently improves performance across both datasets for all ConvRNN variants, significantly reducing prediction errors (MSE and MAE) and enhancing structural similarity (SSIM), without altering the backbone architecture. This clearly shows that GFM, as a plug-and-play module, can stably and effectively enhance the predictive accuracy and robustness of ConvRNN models in different scenarios.

To evaluate the role of the MGM module in MotionRNN, we compare MotionRNN models employing MotionGRU and those incorporating the MGM module on the CIKM2017, Shanghai2020, and TaxiBJ datasets, with the results shown in Table \ref{tab:motionrnn_mgm}. It can be seen that across all datasets, MotionRNN with the MGM module achieves lower MSE and MAE and higher SSIM, exhibiting consistent and significant performance improvements. Notably, the MGM module is inspired by the modeling of non-rigid deformations, enabling it to effectively capture the growth, dissipation, and deformation of targets during motion. This allows the model to demonstrate stronger adaptability and generalization in complex spatiotemporal dynamics. These results indicate that MGM is effective not only for typical non-stationary sequences such as meteorological and traffic data but also, as a plug-and-play module, holds broader application prospects, further validating the rationality and practical value of its design.

\begin{table*}[ht]
\huge
\centering
\caption{Ablation study results on the CIKM2017 dataset. A checkmark (\checkmark) indicates the presence of a module, while a cross (\XSolidBrush) indicates its absence. "Simplified" denotes a lightweight variant of GFM.}
\label{tab:seven}
\resizebox{0.80\textwidth}{!}{%
\begin{tabular}{@{}l|ccc|cccccc@{}}
\toprule
\textbf{Model index} & \textbf{GFM} & \textbf{SAM} & \textbf{MGM} & \textbf{Params} & \textbf{FLOPs} & \textbf{FPS} & \textbf{MSE (↓)} & \textbf{MAE (↓)} & \textbf{SSIM (↑)} \\ \midrule
PredRNN           & \XSolidBrush          & \XSolidBrush            & \XSolidBrush       & 23.8M  & 85.43G   &  58.63  & 28.8126                 & 159.404          & 0.7303            \\
MotionRNN         & \XSolidBrush            & \XSolidBrush            & MotionGRU   & 26.9M  & 94.80G &  59.57  & 27.2091                 & 155.827          & 0.7406            \\ \hline
(A)               & \XSolidBrush            & \XSolidBrush            & \checkmark & 26.9M  & 94.80G &  57.37  & 26.4196                 & 153.801          & 0.7462            \\
(B)               & \XSolidBrush            & \checkmark   & \XSolidBrush       & 31.4M  & 99.67G   & 55.65  & 26.9540                 & 155.525          & 0.7444            \\
(C)               & \checkmark   & \XSolidBrush            & \XSolidBrush      & 30.4M  & 94.80G & 56.02  & 26.7479                 & 155.658          & 0.7448            \\
(D)               & \checkmark   & \checkmark   & \XSolidBrush       & 31.4M  & 0.112T  &  52.57   & \textcolor{blue}{\underline{25.2383}}                 & 151.723          & \textbf{\textcolor{red}{0.7528}}            \\ \hline
\textbf{GMG-s (Ours)}            & Simplified   & \checkmark   & \checkmark  & 29.2M  & 0.104T &  52.21  & 25.3616                 & \textcolor{blue}{\underline{151.334}}          & 0.7501            \\
\textbf{GMG-m (Ours)}               & \checkmark   & \XSolidBrush            & \checkmark  & 30.4M  & 0.107T & 55.67   & 26.1861                 & 152.472          & 0.7487            \\
\textbf{GMG-L (Ours)}            & \checkmark   & \checkmark   & \checkmark & 31.44M & 0.112T &  49.11  & \textbf{\textcolor{red}{25.0215}}        & \textbf{\textcolor{red}{149.493}} & \textcolor{blue}{\underline{0.7513}}   \\ \bottomrule
\end{tabular}%
}
\end{table*}

\begin{table}[htbp]
  \centering
  \caption{Ablation on GMG Layer Number on Shanghai2020.}
  \label{tab:eight}
  \begin{tabular}{ccccccc}
    \toprule
    \textbf{Layer} & \textbf{Params} & \textbf{FLOPs} & \textbf{FPS} & \textbf{MSE} & \textbf{MAE} & \textbf{SSIM} \\
    \midrule
    2 & 13.67M & 66.524G & 65.59 & 5.8221 & 44.498 & 0.9033 \\
    3 & 22.58M & 0.110T  & 38.57 & 4.9531 & 39.828 & 0.9165 \\
    4 & 31.44M & 0.152T  & 27.54 & 4.0308 & 35.771 & 0.9300 \\
    5 & 40.41M & 0.196T  & 21.17 & 4.3757 & 36.300 & 0.9260 \\
    \bottomrule
  \end{tabular}
\end{table}

\begin{table}[ht]
	\centering
	\small
	\caption{Ablation study of the Global Focus Module (GFM) on CIKM2017 dataset.}
	\label{tab:gfm_ablation}
	\resizebox{0.48\textwidth}{!}{%
		\begin{tabular}{@{}l|ccc@{}}
			\toprule
			\textbf{Model}    & \textbf{MSE (↓)} & \textbf{MAE (↓)} & \textbf{SSIM (↑)} \\ \midrule
			ConvLSTM          & 28.8162 & 161.242 & 0.7359 \\
			+ GFM             & 26.6154 ($\Delta$-2.2008) & 157.141 ($\Delta$-4.1010) & 0.7436 ($\Delta$+0.0077) \\
			PredRNN           & 28.8126 & 159.404 & 0.7303 \\
			+ GFM             & 26.7479 ($\Delta$-2.0647) & 155.658 ($\Delta$-3.7460) & 0.7448 ($\Delta$+0.0145) \\
			PredRNNpp         & 28.0987 & 159.258 & 0.7362 \\
			+ GFM             & 26.9204 ($\Delta$-1.1783) & 156.211 ($\Delta$-3.0470) & 0.7438 ($\Delta$+0.0076) \\
			MotionRNN         & 27.2091 & 155.827 & 0.7406 \\
			+ GFM             & 26.1861 ($\Delta$-1.0230) & 152.472 ($\Delta$-3.3550) & 0.7487 ($\Delta$+0.0081) \\
			PredRNNv2         & 28.0085 & 161.508 & 0.7374 \\
			+ GFM             & 27.3749 ($\Delta$-0.6336) & 160.649 ($\Delta$-0.8590) & 0.7401 ($\Delta$+0.0027) \\ 
			\bottomrule
		\end{tabular}%
	}
\end{table}

\begin{table}[ht]
	\centering
	\small
	\caption{Ablation study of the Global Focus Module (GFM) on TaxiBJ dataset.}
	\label{tab:gfm_ablation2}
	\resizebox{0.48\textwidth}{!}{%
		\begin{tabular}{@{}l|ccc@{}}
			\toprule
			\textbf{Model}    & \textbf{MSE$\times$100 (↓)} & \textbf{MAE (↓)} & \textbf{SSIM (↑)} \\ \midrule
			ConvLSTM          & 40.0678 & 16.1892 & 0.9819 \\
			+ GFM             & 35.2095 ($\Delta$-4.8583) & 15.2149 ($\Delta$-0.9743) & 0.9842 ($\Delta$+0.0023) \\
			PredRNN           & 35.0348 & 15.1302 & 0.9844 \\
			+ GFM             & 32.8259 ($\Delta$-2.2089) & 14.9933 ($\Delta$-0.1369) & 0.9846 ($\Delta$+0.0002) \\
			PredRNNpp         & 41.8227 & 15.9766 & 0.9824 \\
			+ GFM             & 32.3835 ($\Delta$-9.4392) & 15.2241 ($\Delta$-0.7525) & 0.9842 ($\Delta$+0.0018) \\
			MotionRNN         & 37.6517 & 16.0009 & 0.9825 \\
			+ GFM             & 34.9205 ($\Delta$-2.7312) & 15.2291 ($\Delta$-0.7718) & 0.9838 ($\Delta$+0.0013) \\
			PredRNNv2         & 45.2737 & 16.6105 & 0.9807 \\
			+ GFM             & 39.0525 ($\Delta$-6.2212) & 15.3422 ($\Delta$-1.2683) & 0.9833 ($\Delta$+0.0026) \\ 
			\bottomrule
		\end{tabular}%
	}
\end{table}

\begin{table}[ht]
	\centering
	\small
	\caption{Comparison of MotionRNN with MotionGRU and Motion Guided Module (MGM) across three datasets.}
	\label{tab:motionrnn_mgm}
	\resizebox{0.5\textwidth}{!}{%
		\begin{tabular}{@{}l|l|ccc@{}}
			\toprule
			\textbf{Dataset} & \textbf{Model} & \textbf{MSE (↓)} & \textbf{MAE (↓)} & \textbf{SSIM (↑)} \\ \midrule
			CIKM2017         & MotionRNN & 27.2091 & 155.827 & 0.7406 \\
			& +MGM & 26.4060 ($\Delta$-0.8031) & 154.574 ($\Delta$-1.253) & 0.7445 ($\Delta$+0.0039) \\ \midrule
			Shanghai2020     & MotionRNN & 4.5867  & 37.871  & 0.9221 \\
			& +MGM & 4.4530  ($\Delta$-0.1337) & 36.971 ($\Delta$-0.900) & 0.9246 ($\Delta$+0.0025) \\ \midrule
			TaxiBJ           & MotionRNN & 0.3765  & 16.000  & 0.9825 \\
			& +MGM & 0.3408  ($\Delta$-0.0357) & 15.297 ($\Delta$-0.703) & 0.9840 ($\Delta$+0.0015) \\ 
			\bottomrule
		\end{tabular}%
	}
\end{table}

\clearpage
\clearpage
\section*{Coclusion}
In this paper, we propose GMG, a video prediction model that integrates Global Vision and Motion Guidance to effectively capture spatial-temporal correlations. By introducing the Global Focus Module (GFM) and Motion Guided Module (MGM), our model generalizes the non-rigid motion deformation process using a balancing factor \( \bm{\alpha}\) and a decay factor \( \bm{\beta}\), enabling more precise predictions.

Extensive experiments on six benchmark datasets demonstrate that GMG significantly enhances prediction performance, surpassing existing mainstream methods while exhibiting strong generalization and application potential across different tasks.

Additionally, we provide model parameter sizes and training durations to facilitate real-world deployment. The lightweight version, GMG-s, also achieves competitive results, highlighting its potential for efficiency-critical applications. Overall, GMG outperforms existing CNN, RNN, and ViT-based models in video prediction, offering a novel solution for complex sequence forecasting.

\begin{IEEEbiography}[{\includegraphics[width=1in,height=1.25in,clip,keepaspectratio]{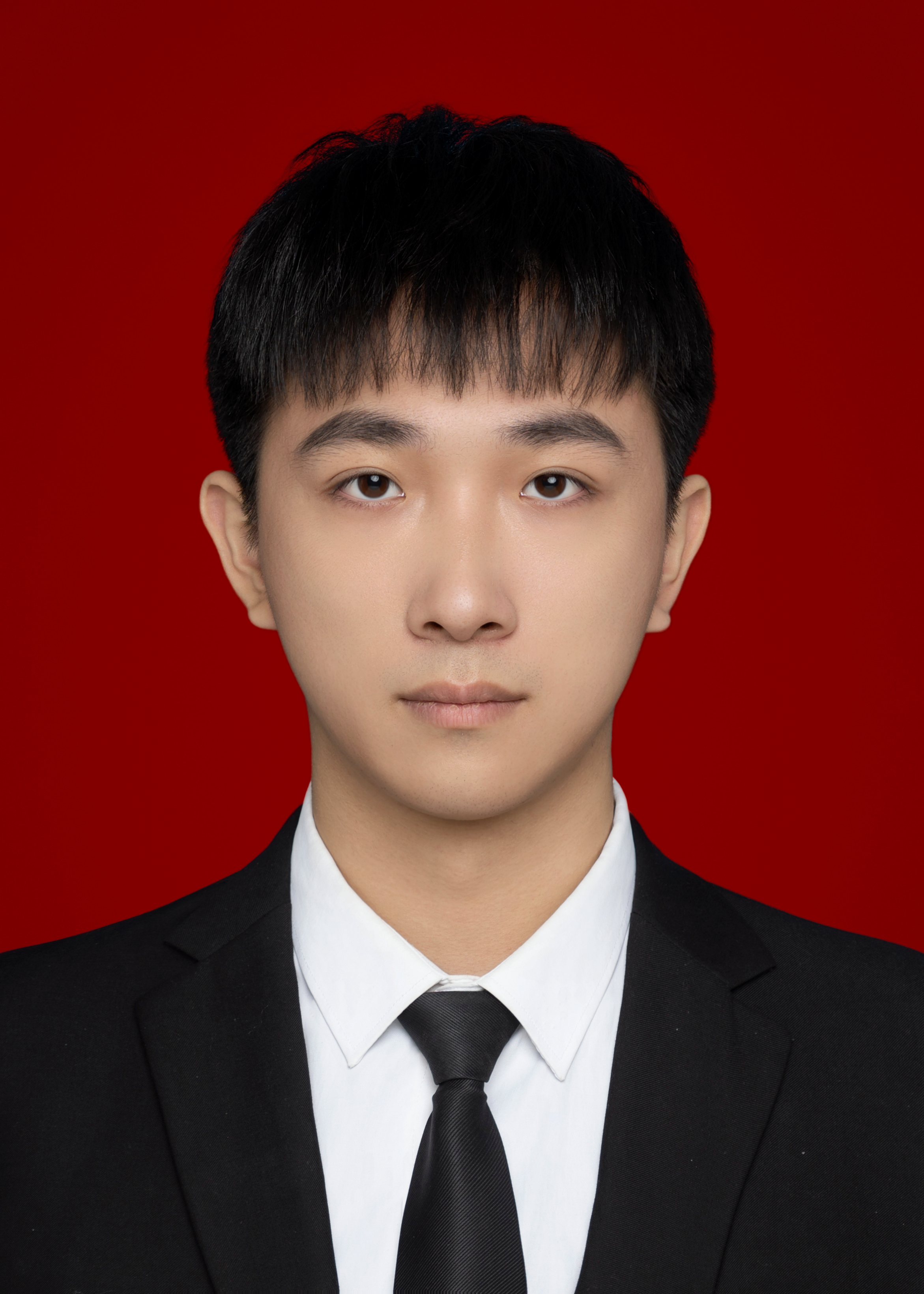}}]{Yuhao Du} graduated with a Bachelor of Science in Information and Computing Science from Central South Forestry University of Science and Technology in 2024. He is pursuing his master's degree in the School of Atmospheric Sciences at Lanzhou University. His research interests are deep learning, machine learning, video prediction, spatio-temporal prediction, and AI for weather prediction.
\end{IEEEbiography}

\begin{IEEEbiography}[{\includegraphics[width=1in,height=1.25in,clip,keepaspectratio]{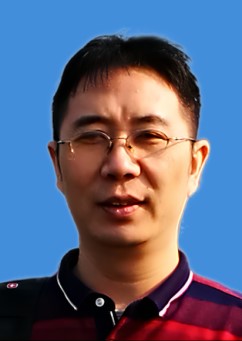}}]{Hui Liu} is currently an Associate Professor at the College of Computer and Mathematics, Central South University of Forestry and Technology, China.He received the B.S. degree in Applied Mathematics from Southeast University, Nanjing, China, in 2001 and the M.S. degree in Forestry Information Engineering from Central South University of Forestry and Technology, Changsha, China, in 2009. He is interested in the areas of computer vision, computer graphics, data analysis and machine learning.
\end{IEEEbiography}

\begin{IEEEbiography}[{\includegraphics[width=1in,height=1.25in,clip,keepaspectratio]{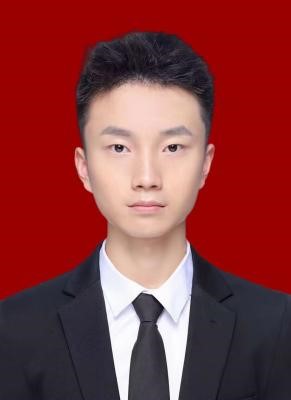}}]{HaoXiang Peng} is pursuing his Bachelor’s degree in Information and Computing Science from Central South Forestry University of Science and Technology. His research interests are deep learning, machine learning, video prediction, and nature language processing.
\end{IEEEbiography}

\begin{IEEEbiography}[{\includegraphics[width=1in,height=1.25in,clip,keepaspectratio]{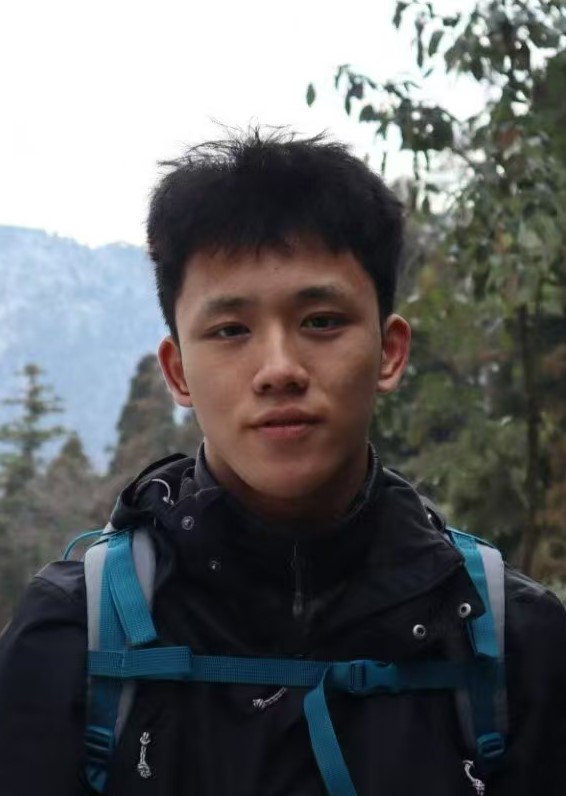}}]{Xinyuan Cheng} received his Bachelor’s degree in German Studies from Shanghai Jiao Tong University in 2024. He is currently pursuing a Master's degree in Computational Linguistics at Ludwig Maximilian University of Munich. His research interests lie in deep learning, machine learning, the evaluation of large language models (LLMs), and the interpretability of LLMs.
\end{IEEEbiography}

\begin{IEEEbiography}[{\includegraphics[width=1in,height=1.25in,clip,keepaspectratio]{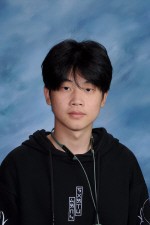}}]{Chengrong Wu} is pursuing his Bachelor’s degree in the department of computer science from the University of Manchester. His research interests are deep learning, machine learning, video prediction, spatiotemporal prediction, AI for weather prediction.
\end{IEEEbiography}

\begin{IEEEbiography}[{\includegraphics[width=1in,height=1.25in,clip,keepaspectratio]{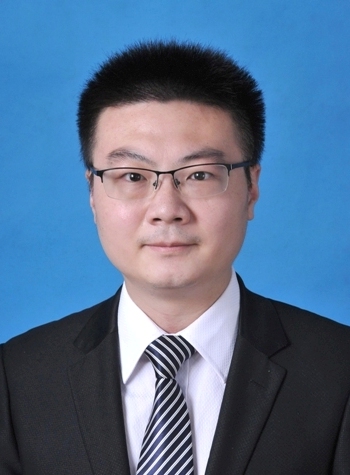}}]{Jiankai Zhang} is a Professor of College of Atmospheric Sciences at Lanzhou University. His research interests include stratosphere-troposphere coupling and impacts of Arctic climate change on midlatitude weather and climate.
\end{IEEEbiography}

\vfill

\end{document}